\UseRawInputEncoding
\documentclass[10pt,twocolumn]{article} 
\pdfoutput=1
\usepackage{simpleConference}
\usepackage{times}
\usepackage{graphicx}
\usepackage{amsmath}
\usepackage{amssymb}
\usepackage{booktabs}
\usepackage{algorithm}
\usepackage{algorithmic}
\usepackage{makecell}
\usepackage{cuted}
\usepackage{float}
\usepackage{capt-of}
\usepackage{hyperref}
\hypersetup{pagebackref,breaklinks,colorlinks}
\usepackage{caption}
\begin{document}

\title{DomainStudio: Fine-Tuning Diffusion Models for Domain-Driven Image Generation using Limited Data}

\author{
  Jingyuan Zhu \\
  Tsinghua University\\
  \texttt{jy-zhu20@mails.tsinghua.edu.cn} \\
   \And
    Huimin Ma \\
    University of Science and Technology Beijing\\
    \texttt{mhmpub@ustb.edu.cn}  \\
    \and
    Jiansheng Chen \\
    University of Science and Technology Beijing \\
    \texttt{jschen@ustb.edu.cn}  \\
    \and
    Jian Yuan  \\
    Tsinghua University\\
    \texttt{jyuan@tsinghua.edu.cn}  \\
}



\maketitle
\thispagestyle{empty}

\begin{strip}
\centering
\includegraphics[width=1.0\linewidth]{ 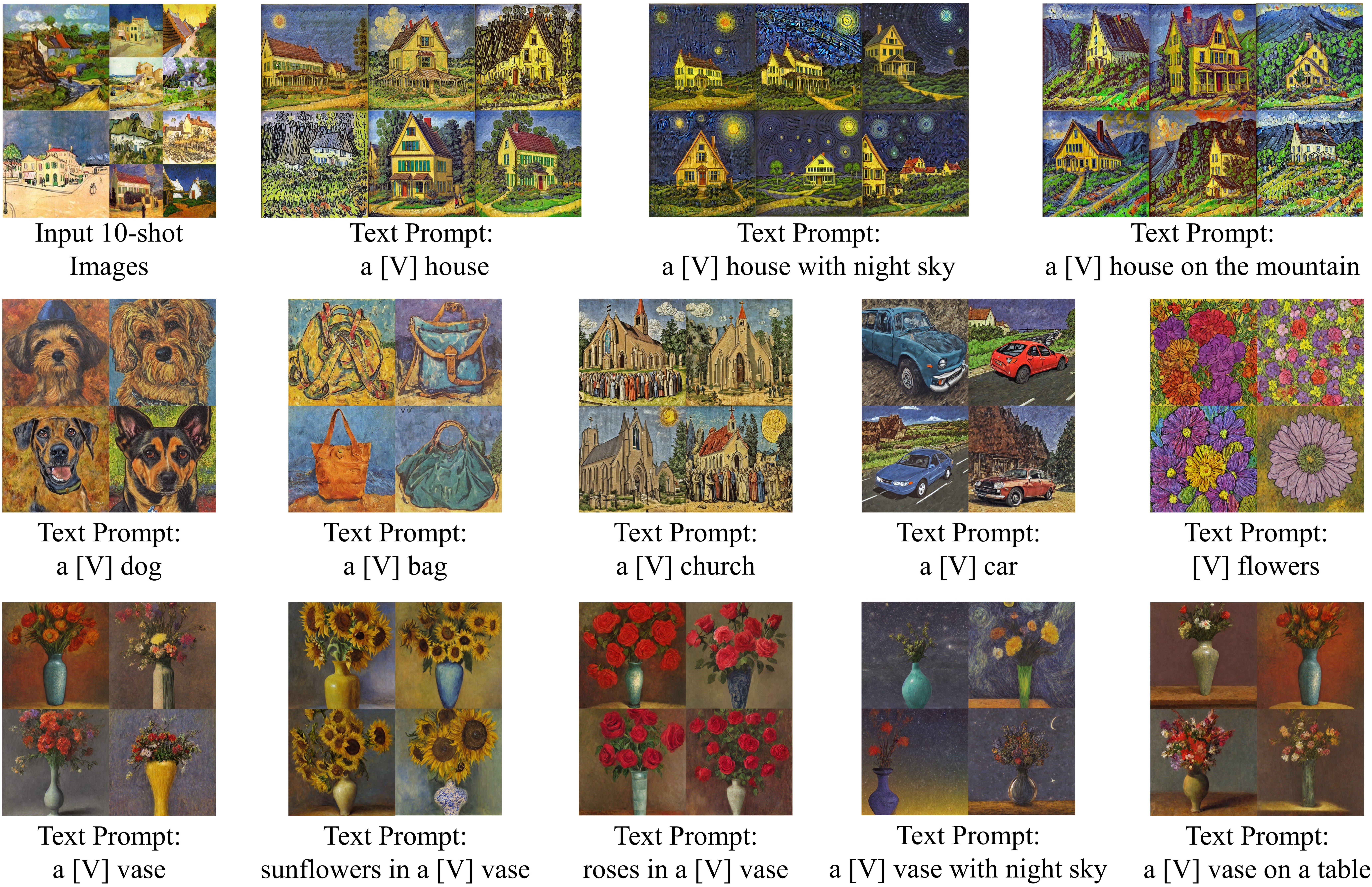}
\captionof{figure}{Given few-shot training samples, the proposed AI image creation approach named DomainStudio can generate: 1) samples containing the same subject as training samples with different contexts (the first row), 2) samples containing subjects different from training samples (the second row), 3) samples containing subjects different from training samples with different contexts (the third row).
\label{vangoghhouse}}
\end{strip}

\begin{abstract}
   Denoising diffusion probabilistic models (DDPMs) have been proven capable of synthesizing high-quality images with remarkable diversity when trained on large amounts of data. Typical diffusion models and modern large-scale conditional generative models like text-to-image generative models are vulnerable to overfitting when fine-tuned on extremely limited data. Existing works have explored subject-driven generation using a reference set containing a few images. However, few prior works explore DDPM-based domain-driven generation, which aims to learn the common features of target domains while maintaining diversity. This paper proposes a novel DomainStudio approach to adapt DDPMs pre-trained on large-scale source datasets to target domains using limited data. It is designed to keep the diversity of subjects provided by source domains and get high-quality and diverse adapted samples in target domains. We propose to keep the relative distances between adapted samples to achieve considerable generation diversity. In addition, we further enhance the learning of high-frequency details for better generation quality. Our approach is compatible with both unconditional and conditional diffusion models. This work makes the first attempt to realize unconditional few-shot image generation with diffusion models, achieving better quality and greater diversity than current state-of-the-art GAN-based approaches. Moreover, this work also significantly relieves overfitting for conditional generation and realizes high-quality domain-driven generation, further expanding the applicable scenarios of modern large-scale text-to-image models.
\end{abstract}

\section{Introduction}
\label{sec:intro}
 Recent advances in generative models including GANs \cite{NIPS2014_5ca3e9b1, DBLP:conf/iclr/BrockDS19, Karras_2019_CVPR, Karras_2020_CVPR, Karras2021}, variational autoencoders (VAEs) \cite{kingma2013auto, rezende2014stochastic, vahdat2020nvae}, and autoregressive models \cite{van2016conditional,chen2018pixelsnail,henighan2020scaling} have realized high-quality image generation with great diversity. Diffusion probabilistic models \cite{sohl2015deep} are introduced to match data distributions by learning to reverse multi-step noising processes. Ho et al. \cite{NEURIPS2020_4c5bcfec} demonstrate the capability of DDPMs to produce high-quality results. Following works \cite{song2020improved, dhariwal2021diffusion, nichol2021improved,kingma2021variational} further optimize the noise addition schedules, network architectures, and optimization targets of DDPMs.  DDPMs show excellent generation results competitive with GANs \cite{Karras_2020_CVPR, DBLP:conf/iclr/BrockDS19} on datasets including CIFAR-10 \cite{krizhevsky2009learning}, LSUN \cite{yu2015lsun}, and ImageNet \cite{van2016conditional}. Moreover, DDPMs have also achieved compelling results in generating videos \cite{ho2022video, harvey2022flexible, yang2022diffusion, zhang2022motiondiffuse}, audios \cite{kong2020diffwave, austin2021structured}, point clouds \cite{zhou20213d, luo2021diffusion, lyu2021conditional, liu2022let}, and biological structures \cite{xu2022geodiff,hoogeboom2022equivariant,geossl}.
 
 Modern DDPMs depend on large amounts of data to train the millions of parameters in their networks like other generative models, which tend to overfit seriously and fail to produce high-quality images with considerable diversity when training data is limited. Unfortunately, it is not always possible to obtain abundant data under some circumstances. A series of GAN-based approaches \cite{wang2018transferring, ada,mo2020freeze, wang2020minegan, ewc, ojha2021few-shot-gan, zhao2022closer} have been proposed to adapt models pre-trained on large-scale source datasets to target datasets using a few available training samples (e.g., 10 images). These approaches utilize knowledge from source models to relieve overfitting but can only achieve limited quality and diversity. Unconditional few-shot image generation with diffusion models remains to be investigated. 
 
 Few-shot text-to-image generation methods \cite{gal2022textual,ruiz2022dreambooth,kumari2023multi,gu2023mixofshow} are designed to preserve the key features of subjects and synthesize the subject with novel scenes, poses, and views utilizing the prior knowledge of large-scale text-to-image models \cite{rombach2021highresolution}. However, it remains challenging to extract the common features of limited target data and realize few-shot generation with diverse subjects instead of producing novel scenes of a certain subject. 

 This work concentrates on domain-driven generation with diffusion models using limited data. More specifically, we aim to adapt diverse source samples containing various subjects to target domains given a few reference images. For example, given only a few house paintings of Van Gogh, we train diffusion models to synthesize paintings of various subjects in the same style. Similarly, we can synthesize diverse baby images using a source model pre-trained on face datasets and a few baby images as training data. 

 We first evaluate the performance of DDPMs fine-tuned on limited data directly and show that both unconditional DDPMs and modern conditional text-to-image DDPMs suffer overfitting problems. Firstly, directly fine-tuned models tend to replicate parts of training samples and cannot learn the common features of limited data, resulting in limited diversity and unreasonable results. In addition, it's sometimes hard for the fine-tuned models to preserve rich details, leading to too smooth generated samples and degraded quality. To this end, we introduce the DomainStudio approach, which keeps the relative distance between generated samples and realizes high-frequency details enhancement during domain adaptation to achieve few-shot and domain-driven generation with remarkable quality and diversity.

 The main contributions of our work are concluded as:

\begin{itemize}
    \item We make the first attempt to evaluate the performance of DDPMs trained from scratch as training data become scarce and further take a closer look at DDPMs fine-tuned on extremely limited data. 
    \item We propose a pairwise similarity loss to keep the relative pairwise distances between generated samples during DDPM domain adaptation for greater diversity. 
    \item We design a high-frequency details enhancement approach from two perspectives, including preserving details provided by source models and learning more details from limited data during DDPM domain adaptation for finer quality.
    \item  We demonstrate the effectiveness of DomainStudio qualitatively and quantitatively on a series of few-shot image generation tasks and show that DomainStudio achieves better generation quality and diversity than current state-of-the-art unconditional GAN-based approaches and conditional DDPM-based text-to-image approaches. 
\end{itemize}

\section{Related Work}
\subsection{Diffusion Models}
DDPMs \cite{sohl2015deep} define a forward noising (diffusion) process adding Gaussian noises $\epsilon$ to training samples $x_0$ and employ a UNet-based neural network $\epsilon_{\theta}$ to approximate the reverse distribution, which can be trained to predict the added noises or the denoised images. Ho et al. \cite{NEURIPS2020_4c5bcfec} demonstrate that predicting $\epsilon$ performs well and achieves high-quality results using a reweighted loss function:
\begin{align}
\label{loss_simple}
    \mathcal{L}_{simple}=E_{t,x_0,\epsilon}\left[||\epsilon-\epsilon_{\theta}(x_t,t)||\right]^2,
\end{align}
where t and $x_t$ represent the diffusion step and corresponding noised image. DDPMs have achieved competitive unconditional generation results on typical large-scale datasets \cite{krizhevsky2009learning,yu2015lsun,van2016conditional}. Besides, classifier guidance is added to realize DDPM-based conditional image generation \cite{dhariwal2021diffusion}. Latent diffusion models \cite{rombach2021highresolution} employ pre-trained autoencoders to compress images into the latent space and achieve high-quality conditional generation using inputs such as text, images, and semantic maps.

\textbf{Text-to-image Generation} Text-driven image generation \cite{crowson2022vqgan,ding2021cogview,gafni2022make,jain2022zero,hinz2020semantic,li2019object,li2019controllable,qiao2019learn,qiao2019mirrorgan,ramesh2021zero,tao2020df,zhang2018photographic} has achieved great success based on GANs \cite{DBLP:conf/iclr/BrockDS19,Karras_2019_CVPR,Karras_2020_CVPR,Karras2021}, transformers \cite{vaswani2017attention}, and diffusion models \cite{NEURIPS2020_4c5bcfec} with the help of image-text representations like CLIP \cite{radford2021learning}. Large-scale text-to-image generative models including Imagen \cite{saharia2022photorealistic}, Parti \cite{yu2022scaling}, CogView2 \cite{ding2022cogview2}, DALL-E2 \cite{ramesh2022hierarchical}, and Stable Diffusion \cite{rombach2021highresolution} further expands application scenarios and improve generation quality. Light-weight fine-tuning methods like Textual Inversion \cite{gal2022textual} and DreamBooth \cite{ruiz2022dreambooth} realize personalization of text-to-image diffusion models. However, these methods are still vulnerable to overfitting and mainly focus on subject-driven generation. Recent works \cite{sohn2023learning,sohn2023styledrop} based on MaskGIT \cite{chang2022maskgit} and MUSE \cite{chang2023muse} tackle similar problems of generating images containing different subjects and sharing the same style with training samples. Our approach is designed to realize few-shot and domain-driven generation with diffusion models and achieves compelling results with high quality and great diversity.


\textbf{Applications} DDPMs have already been applied to many aspects of applications such as image super-resolution \cite{li2022srdiff, saharia2022image, rombach2022high,ho2022cascaded}, image translation \cite{saharia2022palette, ozbey2022unsupervised}, semantic segmentation \cite{baranchuk2021label, asiedu2022decoder}, few-shot generation for unseen classes \cite{giannone2022few,sinha2021d2c}, and natural language processing \cite{austin2021structured, li2022diffusion, chen2022analog}. Besides, DDPMs are combined with other generative models including GANs \cite{xiao2021tackling, wang2022diffusion}, VAEs \cite{vahdat2021score, huang2021variational, luo2022understanding}, and autoregressive models \cite{rasul2021autoregressive, hoogeboom2021autoregressive}. Different from existing works, this paper focuses on model-level, unconditional, few-shot image generation with DDPM-based approaches.

\begin{figure*}[t]
    \centering
    \includegraphics[width=1.0\linewidth]{ 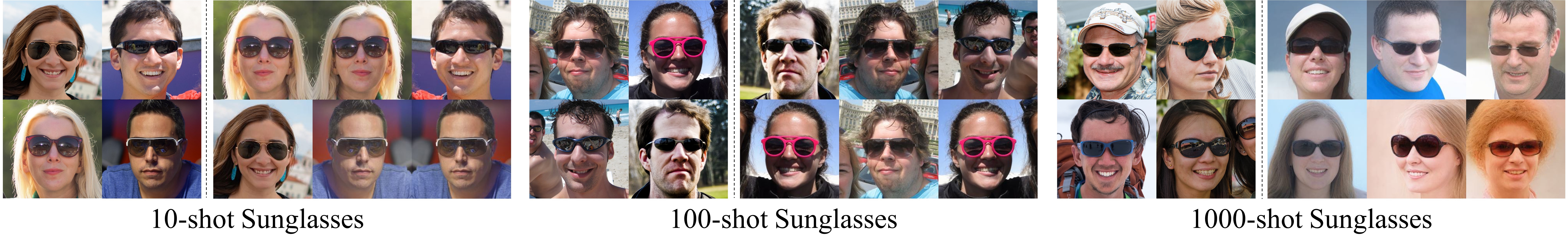}
    \caption{ For small-scale Sunglasses datasets containing 10, 100, and 1000 images, \textbf{Left}: samples picked from the small-scale datasets, \textbf{Right}: samples produced by DDPMs trained on the small-scale datasets from scratch.}
    \label{scratch}
\end{figure*}

\subsection{Few-shot Image Generation}
Few-shot image generation aims to achieve high-quality generation with great diversity using only a few available training samples. However, modern generative models easily overfit and suffer severe diversity degradation when trained on limited data (e.g., 10 images). They tend to replicate training samples instead of generating diverse images following similar distributions. GAN-based few-shot image generation approaches mainly follow TGAN \cite{wang2018transferring} to adapt GANs pre-trained on large source domains, including ImageNet \cite{van2016conditional}, LSUN \cite{yu2015lsun}, and FFHQ \cite{Karras_2020_CVPR}, to related target domains with limited data. Augmentation approaches \cite{tran2021data, zhao2020differentiable, zhao2020image, ada} also help improve generation diversity. BSA \cite{noguchi2019image} fixes all the parameters except for the scale and shift parameters in the generator. FreezeD \cite{mo2020freeze} freezes parameters in high-resolution layers of the discriminator to relieve overfitting. MineGAN \cite{wang2020minegan} uses additional fully connected networks to modify noise inputs for the generator. EWC \cite{ewc} makes use of elastic weight consolidation to regularize the generator by making it harder to change the critical weights which have higher Fisher information \cite{2017A} values. CDC \cite{ojha2021few-shot-gan} proposes a cross-domain consistency loss and patch-level discrimination to build a correspondence between source and target domains. DCL \cite{zhao2022closer} utilizes contrastive learning to push away the generated samples from real images and maximize the similarity between corresponding image pairs in source and target domains. The proposed DomainStudio approach follows similar strategies to adapt models pre-trained on large source domains to target domains. Our experiments show that DDPMs are qualified for few-shot image generation tasks and can achieve better results than current state-of-the-art GAN-based approaches in generation quality and diversity.

\section{DDPMs Trained on Small-scale Datasets}
\label{section3}
\subsection{Training from Scratch}
We first train DDPMs on small-scale datasets containing various numbers of images from scratch. We analyze generation diversity qualitatively and quantitatively to study when do DDPMs overfit as training samples decrease. 

\textbf{Basic Setups} We sample 10, 100, and 1000 images from FFHQ-babies (Babies), FFHQ-sunglasses (Sunglasses) \cite{ojha2021few-shot-gan}, and LSUN Church \cite{yu2015lsun} respectively as small-scale training datasets. The image resolution of all the datasets is set as $256\times 256$. We follow the model setups in prior works \cite{nichol2021improved, dhariwal2021diffusion} used for LSUN $256^2$ \cite{yu2015lsun}. The max diffusion step $T$ is set as 1000. We use a learning rate of 1e-4 and a batch size of 48. We train DDPMs for 40K iterations on datasets containing 10 or 100 images and 60K iterations on datasets containing 1000 images empirically.

\textbf{Qualitative Evaluation} In Fig. \ref{scratch}, we visualize the generated samples of DDPMs trained from scratch on few-shot Sunglasses datasets and provide some training samples for comparison (more generated and training samples are added in Appendix \ref{appendix_scratch}). We observe that DDPMs overfit and tend to replicate training samples when datasets are limited to 10 or 100 images. Since some training samples are flipped in the training process as a step of data augmentation, we can also find some generated images symmetric to the training samples. While for datasets containing 1000 images, DDPMs can generate diverse samples following similar distributions of training samples instead of replicating them. The overfitting problem is relatively alleviated. However, the generated samples are coarse and lack high-frequency details compared with training samples.

\begin{figure*}[t]
    \centering
    \includegraphics[width=1.0\linewidth]{ 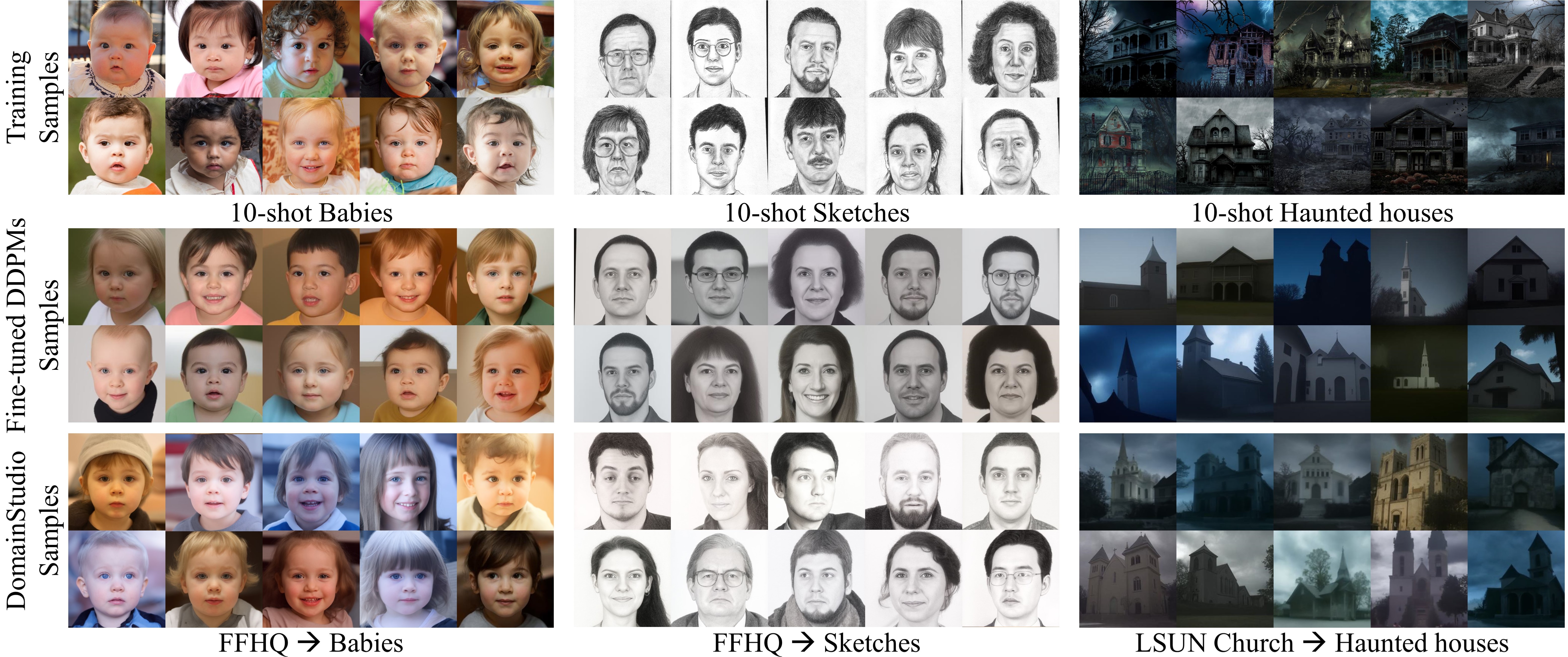}
    \caption{DDPM-based image generation samples on 10-shot FFHQ $\rightarrow$ Babies, FFHQ $\rightarrow$ Sketches, and LSUN Church $\rightarrow$ Haunted houses.}
    \label{result1}
\end{figure*}

\begin{table}[tbp]
\centering
\begin{tabular}{c|c|c|c}
Number of Samples & Babies & Sunglasses & Church \\
\hline
$10$ & $0.2875$ & $0.3030$ & $0.3136$ \\
$100$ & $0.3152$ & $0.3310$ & $0.3327$ \\
$1000$  & $\pmb{0.4658}$ & $\pmb{0.4819}$ & $\pmb{0.5707}$ \\
\hline
$10$ (+ flip) & $0.1206$ & $0.1217$ & $0.0445$\\
$100$ (+ flip) & $0.1556$ & $0.1297$ & $0.1177$ \\
$1000$ (+ flip) & $\pmb{0.4611}$ & $\pmb{0.4726}$ & $\pmb{0.5625}$ \\
\end{tabular}
\caption{Nearest-LPIPS ($\uparrow$) results of DDPMs trained from scratch on several small-scale datasets.}
\label{nearlpips}
\end{table}

\textbf{Quantitative Evaluation} LPIPS \cite{zhang2018unreasonable} is proposed to evaluate the perceptual distances \cite{johnson2016perceptual} between images. We propose a Nearest-LPIPS metric based on LPIPS to evaluate the generation diversity of DDPMs trained on small-scale datasets. More specifically, we first generate 1000 images randomly and find the most similar training sample having the lowest LPIPS distance to each generated sample. Nearest-LPIPS is defined as the LPIPS distances between generated samples and the most similar training samples in correspondence averaged over all the generated samples. If a generative model reproduces the training samples exactly, the Nearest-LPIPS metric will have a score of zero. Larger Nearest-LPIPS values indicate lower replication rates and greater diversity relative to training samples.

\begin{figure}[t]
    \centering
    \includegraphics[width=1.0\linewidth]{ 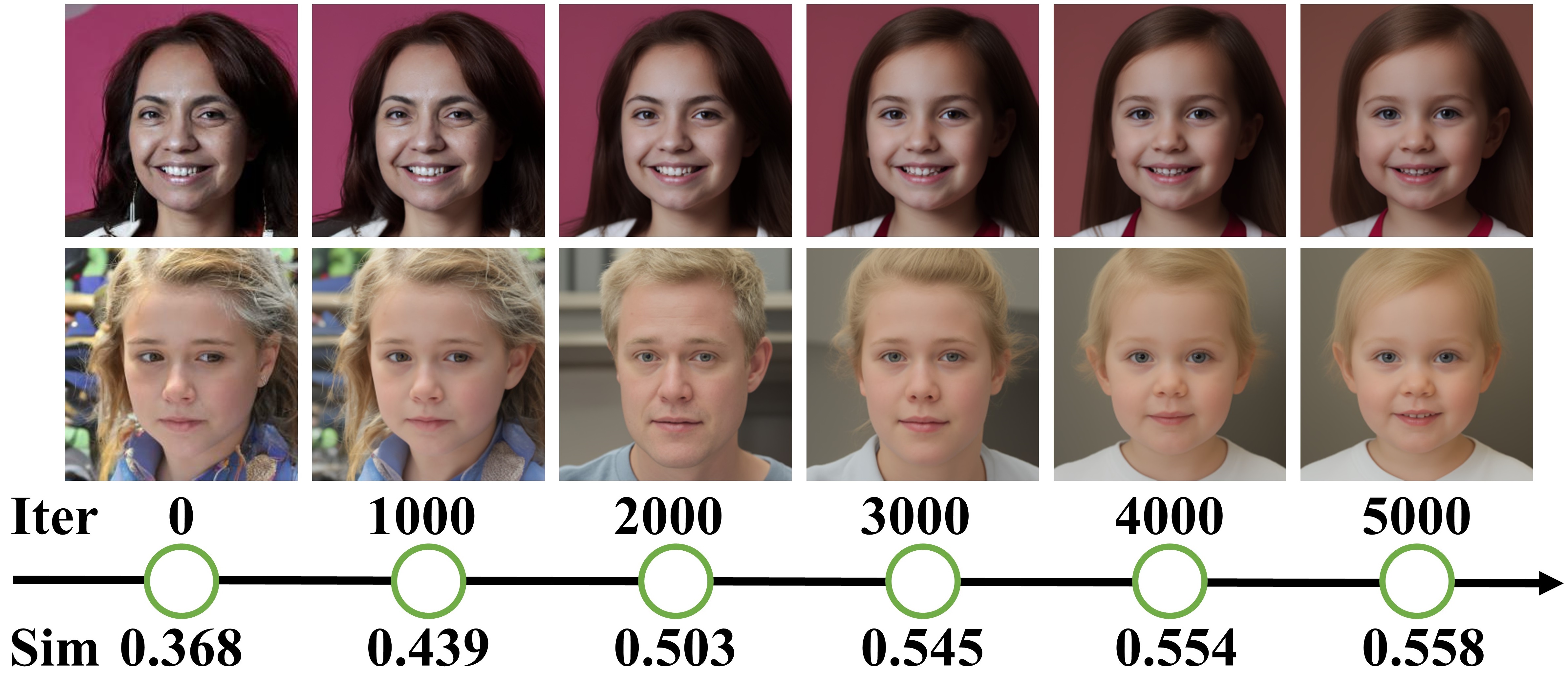}
    \caption{Two samples synthesized from fixed noise inputs by the directly fine-tuned DDPM on 10-shot FFHQ $\rightarrow$ Babies become more and more similar throughout training, as shown by the increasing cosine similarity.}
    \label{degrade}
\end{figure}

We provide the Nearest-LPIPS results of DDPMs trained from scratch on small-scale datasets in the top part of Table \ref{nearlpips}. For datasets containing 10 or 100 images, we have lower Nearest-LPIPS values. While for datasets containing 1000 images, we get measurably improved Nearest-LPIPS values. To avoid the influence of generated images symmetric to training samples, we flip all the training samples as supplements to the original datasets and recalculate the Nearest-LPIPS metric. The results are listed in the bottom part of Table \ref{nearlpips}. With the addition of flipped training samples, we find apparently lower Nearest-LPIPS values for datasets containing 10 or 100 images. However, we get almost the same Nearest-LPIPS results for DDPMs trained on larger datasets containing 1000 images, indicating that these models can generate diverse samples different from the original or symmetric training samples.

To sum up, it becomes harder for DDPMs to learn the representations of datasets as training data become scarce. When trained on limited data from scratch, DDPMs fail to match target data distributions exactly and cannot produce high-quality and diverse samples.

\subsection{Fine-tuning}
Furthermore, we fine-tune DDPMs pre-trained on large-scale source datasets using limited target data directly to take a closer look at the performance of DDPMs trained on limited data. 

The fine-tuned unconditional DDPMs achieve faster convergence to target domains. They only need 3K-4K iterations to converge in our experiments. As shown in the middle row of Fig. \ref{result1}, they can produce diverse results utilizing only 10 training samples. However, the generated samples lack ample high-frequency details and share similar features like hairstyles and facial expressions, leading to the degradation of generation quality and diversity. Compared with pre-trained models, the degradation of fine-tuned models mainly comes from the excessively shortened relative distances between generated samples. As shown in Fig. \ref{degrade}, two samples synthesized from fixed noise inputs by the directly fine-tuned DDPM become increasingly similar (e.g., eyes and facial expressions) throughout training, losing various features and high-frequency details.

\begin{figure}[t]
    \centering
    \includegraphics[width=1.0\linewidth]{ 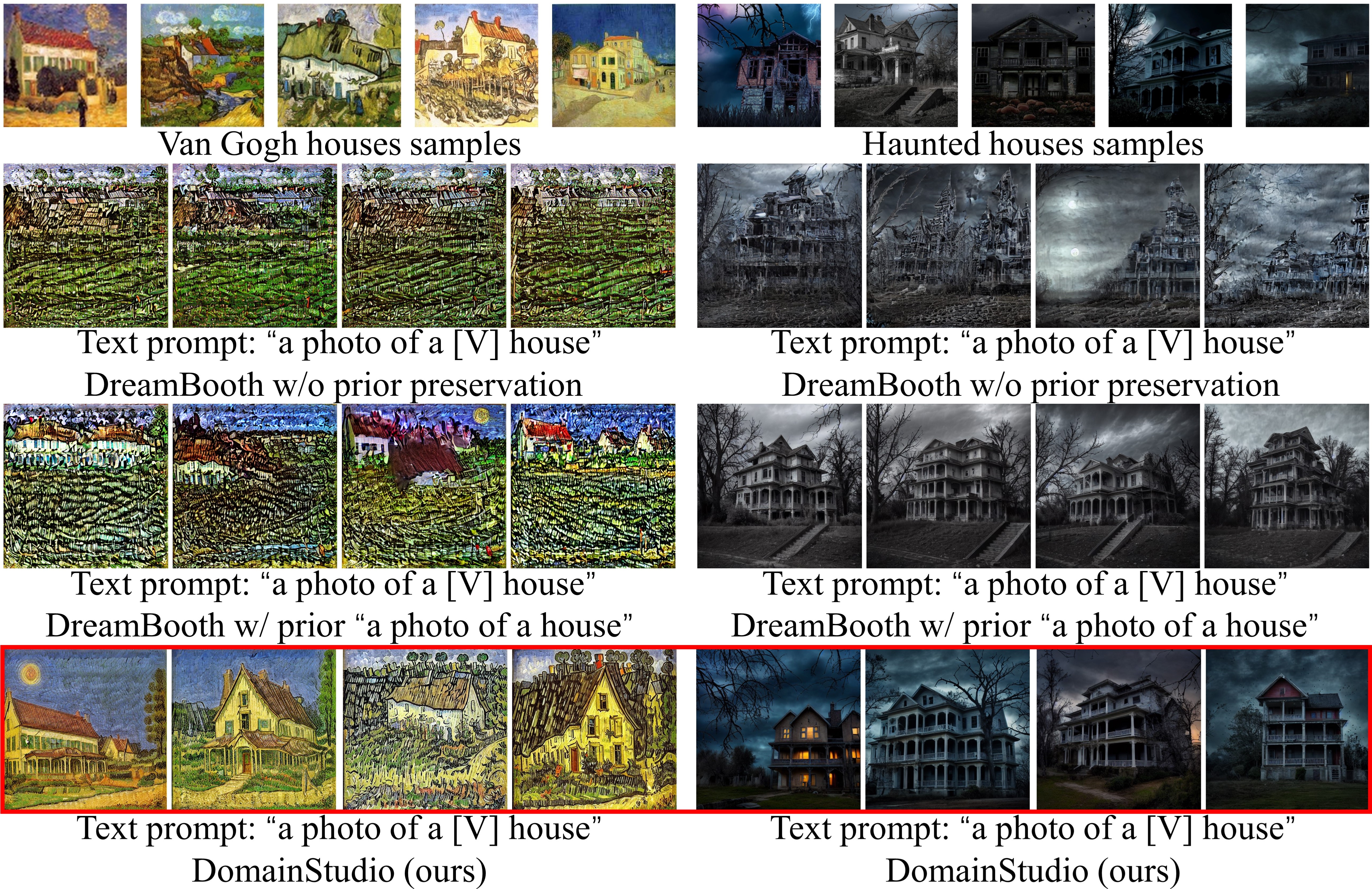}
    \caption{Stable Diffusion v1.4 fine-tuned by DreamBooth and DomainStudio on 10-shot datasets with the same text prompts.}
    \label{degrade3}
\end{figure}

In addition, we fine-tune the large text-to-image generative model Stable Diffusion on 10-shot Van Gogh houses and haunted houses, respectively. We employ a unique identifier [V] to avoid using prior knowledge of target domains in the Stable Diffusion model. As shown in Fig. \ref{degrade3}, DreamBooth without prior preservation severely overfits and produces low-quality samples. The full DreamBooth approach still overfits and gets limited diversity with replicated samples. They tend to preserve the subjects in training samples instead of generating diverse samples following similar distributions, which is pursued by domain-driven generation.

\begin{figure*}[t]
    \centering
    \includegraphics[width=1.0\linewidth]{ 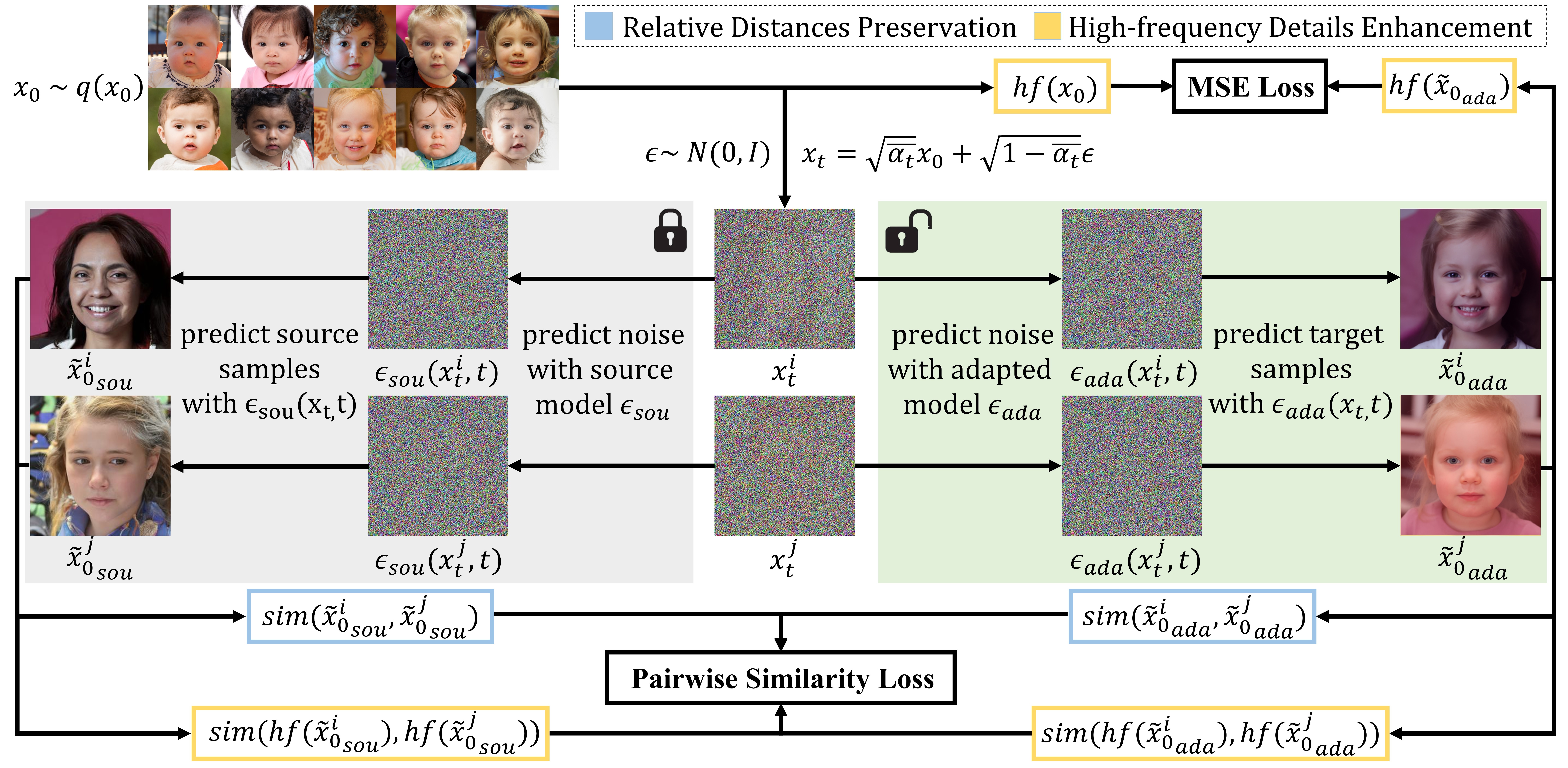}
    \caption{\textbf{Overview of the DomainStudio approach applied to unconditional image generation.} A pairwise similarity loss is introduced to keep the relative pairwise distances of generated samples and their high-frequency details of adapted models similar to source models during adaptation. DomainStudio guides adapted models to learn more high-frequency details from limited data using the MSE loss between high-frequency details extracted from training data and adapted samples. DomainStudio generates high-quality and diverse samples sharing similar styles with few-shot training data.}
    \label{pairwise}
\end{figure*}

\section{Unconditional Image Generation}
In this section, we illustrate the DomainStudio approach on unconditional image generation with typical diffusion models. We propose to regularize the domain adaptation process by keeping the relative pairwise distances between adapted samples similar to source samples (Sec \ref{42}). Besides, we guide adapted models to learn high-frequency details from limited data and preserve high-frequency details learned from source domains (Sec \ref{44}). Our approach fixes source models $\epsilon_{sou}$ as reference for adapted models $\epsilon_{ada}$. The weights of adapted models are initialized to the weights of source models and adapted to target domains. An overview of the proposed DomainStudio approach for unconditional image generation is illustrated in Fig. \ref{pairwise} using 10-shot FFHQ $\rightarrow$ Babies as an example.

\subsection{Relative Distances Preservation}
\label{42}
We design a pairwise similarity loss to maintain the distributions of relative pairwise distances between generated images during domain adaptation. To construct N-way probability distributions for each image, we sample a batch of noised images $\left\lbrace x_t^{n} \right\rbrace_{n=0}^{N}$ by randomly adding Gaussian noises to training samples $x_0\sim q(x_0)$. Then both source and adapted models are applied to predict the fully denoised images $\left\lbrace \tilde{x}_0^{n} \right\rbrace_{n=0}^{N}$. We have the prediction of $\tilde{x}_0$ in terms of $x_t$ and $\epsilon_{\theta}(x_t,t)$ as follows:
\begin{align}
\label{eq11}
    \tilde{x}_0 = \frac{1}{\sqrt{\overline{\alpha}_t}}x_t-\frac{\sqrt{1-\overline{\alpha}_t}}{\sqrt{\overline{\alpha}_t}}\epsilon_{\theta}(x_t,t).
\end{align}
Cosine similarity is employed to measure the relative distances between the predicted images $\tilde{x}_0$. The probability distributions for $\tilde{x}_0^{i}\ (0\leq i \leq N)$ in source and adapted models are as follows:
\begin{align}
    p_{i}^{sou} = sfm(\left\lbrace sim(\tilde{x}_{0_{sou}}^{i},\tilde{x}_{0_{sou}}^{j})\right\rbrace_{\forall i\neq j}), \label{16} \\ 
    p_{i}^{ada} = sfm(\left\lbrace sim(\tilde{x}_{0_{ada}}^{i},\tilde{x}_{0_{ada}}^{j})\right\rbrace_{\forall i\neq j}), \label{17}
\end{align}
where $sim$ and $sfm$ denote cosine similarity and softmax function, respectively. Then we have the pairwise similarity loss for generated images as follows:
\begin{align}
\label{18}
    \mathcal{L}_{img}(\epsilon_{sou},\epsilon_{ada}) = \mathbb{E}_{t,x_0,\epsilon} \sum_{i} D_{KL} (p_{i}^{ada}\,||\, p_{i}^{sou}),
\end{align}
where $D_{KL}$ represents KL-divergence. $\mathcal{L}_{img}$ prevents adapted samples from being too similar to each other or replicating training samples. Instead, adapted models are encouraged to learn the common features of training data and preserve diverse features learned from source domains. 


\subsection{High-frequency Details Enhancement}
\label{44}
To begin with, we employ the typical Haar wavelet transformation \cite{0The} to disentangle images into multiple frequency components. Haar wavelet transformation contains four kernels including $LL^T$, $LH^T$, $HL^T$, $HH^T$, where $L$ and $H$ represent the low and high pass filters, respectively:
\begin{align}
    L^T = \frac{1}{\sqrt{2}}[1,1], \quad H^T = \frac{1}{\sqrt{2}}[-1,1].
\end{align} 
Haar wavelet transformation decomposes inputs into four frequency components $LL$, $LH$, $HL$, and $HH$. $LL$ contains fundamental structures of images while other high-frequency components $LH$, $HL$, and $HH$ contain rich details of images. We define $hf$ as the sum of these high-frequency components needing enhancement:
\begin{align}
    hf = LH+HL+HH.
\end{align}

We implement high-frequency details enhancement from two perspectives. Firstly, we use the proposed pairwise similarity loss to preserve high-frequency details learned from source domains. Similarly, the probability distributions for the high-frequency components of $\tilde{x}_0^{i}\ (0\leq i \leq N)$ in source and adapted models and the pairwise similarity loss for the high-frequency components are as follows:
\begin{small}
\begin{align}
    & pf_{i}^{sou} = sfm(\left\lbrace sim(hf(\tilde{x}_{0_{sou}}^{i}),hf(\tilde{x}_{0_{sou}}^{j}))\right\rbrace_{\forall i\neq j}), \\
    & pf_{i}^{ada} = sfm(\left\lbrace sim(hf(\tilde{x}_{0_{ada}}^{i}),hf(\tilde{x}_{0_{ada}}^{j}))\right\rbrace_{\forall i\neq j}), \\
    & \mathcal{L}_{hf}(\epsilon_{sou},\epsilon_{ada}) = \mathbb{E}_{t,x_0,\epsilon} \sum_{i} D_{KL} (pf_{i}^{ada}\,||\, pf_{i}^{sou}),
\end{align}
\end{small}

Secondly, we guide adapted models to learn more high-frequency details from limited training data by minimizing the mean squared error (MSE) between the high-frequency components in adapted samples $\tilde{x}_{0_{ada}}$ and training samples $x_0$ as follows:
\begin{align}
    \mathcal{L}_{hfmse}(\epsilon_{ada}) = \mathbb{E}_{t,x_0,\epsilon} \left[||hf(\tilde{x}_{0_{ada}})-hf(x_0)||\right]^2.
\end{align}

\subsection{Overall Optimization Target}
The overall optimization target of DomainStudio for unconditional image generation combines all the methods proposed above to realize relative distances preservation and high-frequency details enhancement: 
\begin{align}
\label{loss}
    \mathcal{L} = \mathcal{L}_{simple} + \lambda_1 \mathcal{L}_{vlb} + \lambda_2 \mathcal{L}_{img} +\lambda_3 \mathcal{L}_{hf} + \lambda_4 \mathcal{L}_{hfmse}.
\end{align}
We follow prior works \cite{nichol2021improved} to set $\lambda_1$ as 0.001 to avoid the variational lower bound loss $\mathcal{L}_{vlb}$ from overwhelming $\mathcal{L}_{simple}$.  We empirically find $\lambda_2,\lambda_3$ ranging between 0.1 and 1.0 and $\lambda_4$ ranging between 0.01 and 0.08 to be effective for few-shot adaptation setups. 

\section{Text-to-image Generation}
\begin{figure*}[t]
    \centering
    \includegraphics[width=0.9\linewidth]{ 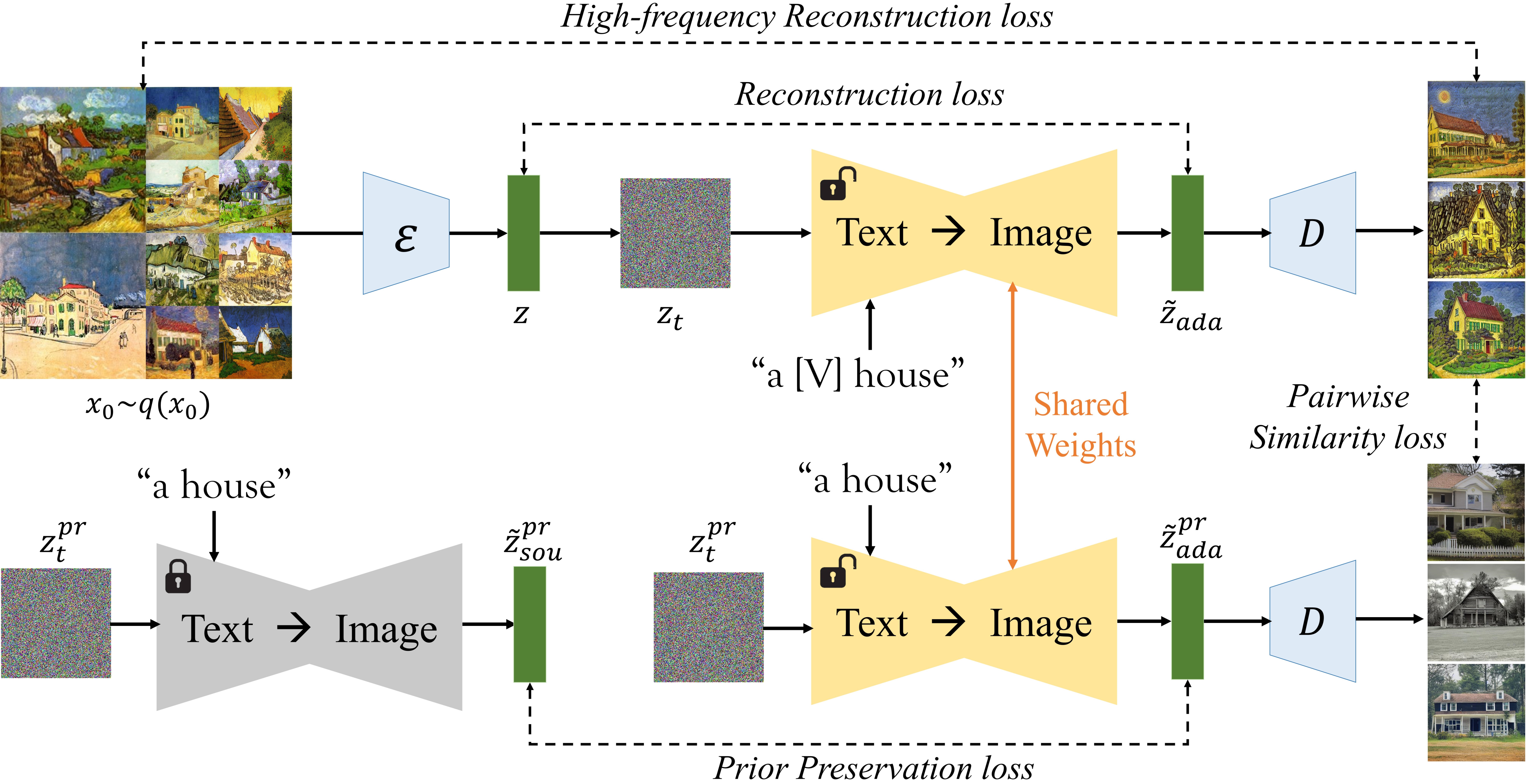}
    \caption{\textbf{Overview of the proposed DomainStudio approach applied to text-to-image diffusion models.} Our approach follows DreamBooth to preserve the information of source domains with prior preservation loss. Similar to unconditional image generation, DomainStudio applies pairwise similarity loss to generated samples and their high-frequency components in source and target domains. In addition, DomainStudio uses high-frequency reconstruction loss to enhance details learning from target domains. }
    \label{pairwise2}
\end{figure*}

Similarly, we propose to realize relative distances preservation (Sec. \ref{51}) and high-frequency details enhancement (Sec. \ref{52}) to realize domain-driven text-to-image generation. An overview of the proposed DomainStudio approach applied to text-to-image generation is illustrated in Fig. \ref{pairwise2}. The pre-trained autoencoders ($\varepsilon+D$) used in text-to-image generation are fixed during fine-tuning. We directly use $\epsilon_{ada}$ and $\epsilon_{sou}$ to represent the pre-trained and adapted text-to-image diffusion models.

Given few-shot training samples and a text prompt ${P}_{tar}$ representing the target domain, we first define a source domain using a text prompt ${P}_{sou}$.  To avoid using the prior knowledge of target domains provided by the large text-to-image model, we employ a unique identifier $\left[V\right]$ to represent the target domains. For example, we define the source domain with the text prompt ``a house" given few-shot house paintings of Van Gogh as training samples which are represented by the text prompt ``a $\left[V\right]$ house". The source and target text prompts ${P}_{sou}$ and $P_{tar}$ are encoded by pre-trained text encoder $\Gamma$ to conditioning vectors $c_{sou}$ and $c_{tar}$.

The adapted model is guided to learn from target samples with the reconstruction function:
\begin{equation}
    \mathcal{L}_{simple} = \mathbb{E}_{t,z,c_{tar},\epsilon}||\epsilon_{ada}(z_t,c_{tar})-\epsilon||^2,
\end{equation}
where $z$ and $z_t$ represent the compressed latent codes of training samples and corresponding noised latent codes. 

DreamBooth \cite{ruiz2022dreambooth} generates source samples $x^{pr}$ with randomly sampled Gaussian noises and the source text condition ${c}_{sou}$ using the pre-trained text-to-image model. Then the pre-trained encoder $\varepsilon$ is employed to compress $x^{pr}$ to latent codes $z^{pr}$. Dreambooth proposes a class-specific prior preservation loss as follows to relieve overfitting for subject-driven generation by preserving the information of source domains:
\begin{equation}
    \mathcal{L}_{pr}=\mathbb{E}_{t,z^{pr},c_{sou},\epsilon^{pr}}||\epsilon_{sou}(z^{pr}_t,c_{sou})-\epsilon_{ada}(z^{pr}_t,c_{sou})||^2,
\end{equation}
where $z^{pr}_t$ represents the source latent codes added with noises $\epsilon^{pr}$.

\subsection{Relative Distances Preservation}
\label{51}
Since the source and target samples have different conditions, it's difficult to build correspondence like unconditional image generation. Therefore, we directly use randomly generated source and adapted samples to keep similar relative distances. Given batches of noised source latent codes $\left\lbrace z_t^{pr,n} \right\rbrace_{n=0}^{N}$ and target latent codes $\left\lbrace z_t^{n} \right\rbrace_{n=0}^{N}$, we build probability distributions for source and adapted samples as follows:
\begin{align}
    p_{i}^{sou} = sfm(\left\lbrace sim(D(\tilde{z}_{ada}^{pr,i}),D(\tilde{z}_{ada}^{pr,j})\right\rbrace_{\forall i\neq j}), \\ 
    p_{i}^{ada} = sfm(\left\lbrace sim(D(\tilde{z}_{ada}^{i}),D(\tilde{z}_{ada}^{j}))\right\rbrace_{\forall i\neq j}),
\end{align}
where $\tilde{z}_{ada}^{pr}$ and $\tilde{z}_{ada}$ represent the denoised latent codes corresponding to source and target samples generated by the adapted model. The pairwise similarity loss for generated images can be expressed as follows:
\begin{align}
    \mathcal{L}_{img}(\epsilon_{ada})=\mathbb{E}_{t,z,z^{pr},\epsilon,\epsilon^{pr}}\sum_{i}D_{KL}(p^{ada}_{i}||p^{sou}_{i}).
\end{align}
$\mathcal{L}_{img}$ encourages the adapted model to keep the diversity of adapted samples similar to source samples. In this way, the adapted model is guided to generate adapted samples containing diverse subjects following target distributions.

\subsection{High-frequency Details Enhancement}
\label{52}
The high-frequency details enhancement for text-to-image generation is also composed of two perspectives. Firstly, the pairwise similarity loss is used to enhance the diversity of high-frequency details in adapted samples. The probability distributions for the high-frequency components of source and adapted samples and the pairwise similarity loss are as follows:
\begin{align}
     pf_{i}^{sou} &= sfm(\left\lbrace sim(hf(D(\tilde{z}_{ada}^{pr,i})),hf(D(\tilde{z}_{ada}^{pr,j}))\right\rbrace_{\forall i\neq j}), \\ 
    pf_{i}^{ada} &= sfm(\left\lbrace sim(hf(D(\tilde{z}_{ada}^{i})),hf(D(\tilde{z}_{ada}^{j})))\right\rbrace_{\forall i\neq j}), \\
    \mathcal{L}_{hf}&(\epsilon_{ada})=\mathbb{E}_{t,z,z^{pr},\epsilon,\epsilon^{pr}}\sum_{i}D_{KL}(pf_{i}^{ada}||pf_{i}^{sou}).
\end{align}

Secondly, we propose the high-frequency reconstruction loss to enhance the learning of high-frequency details from training samples as follows:
\begin{align}
    \mathcal{L}_{hfmse}(\epsilon_{ada})=\mathbb{E}_{t,x_0,z,\epsilon}||hf(D(\tilde{z}_{ada})) - hf(x_0)||^2.
\end{align}

\subsection{Overall Optimization Target}
The overall optimization target of DomainStudio for text-to-image generation combines the proposed methods for relative distances preservation and high-frequency details enhancement with DreamBooth:
\begin{align}
    \mathcal{L} = \mathcal{L}_{simple}+\lambda_1 \mathcal{L}_{pr}+ \lambda_2 \mathcal{L}_{img} +\lambda_3 \mathcal{L}_{hf} + \lambda_4 \mathcal{L}_{hfmse}.
\end{align}
We follow DreamBooth to set $\lambda_1$ as 1 and empirically find $\lambda_2,\lambda_3$ ranging between 1e+2 and 5e+2 and $\lambda_4$ ranging between 0.1 and 1.0 to be effective for most text-to-image adaptation setups.

\begin{figure*}[t]
    \centering
    \includegraphics[width=1.0\linewidth]{ 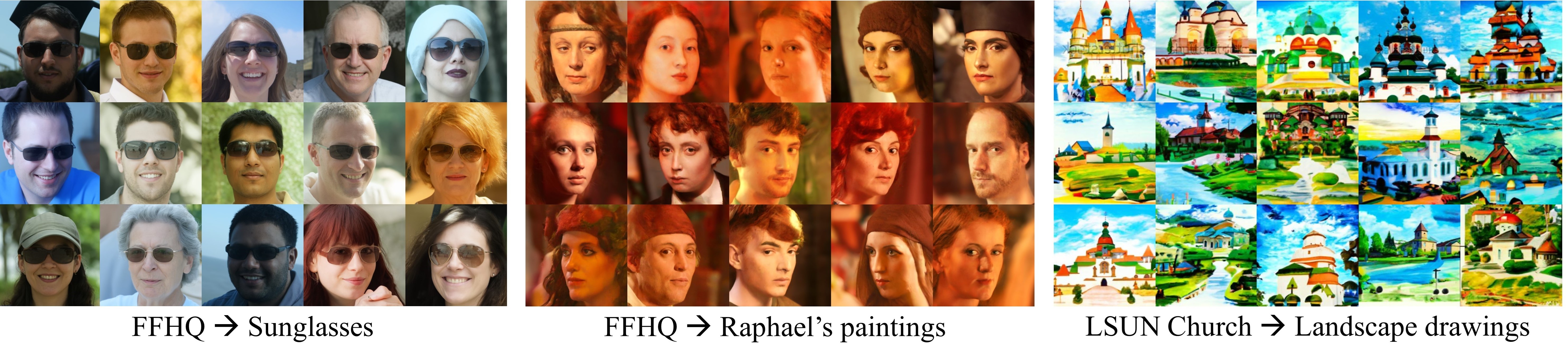}
    \caption{DomainStudio samples on 10-shot FFHQ $\rightarrow$ Sunglasses, FFHQ $\rightarrow$ Raphael's paintings, and LSUN Church $\rightarrow$ Landscape drawings.}
    \label{result2}
\end{figure*}

\section{Experiments}
\label{422}
To demonstrate the effectiveness of DomainStudio, we evaluate it with a series of few-shot fine-tuning tasks using extremely few training samples ($\leq$ 10 images). The performance of DomainStudio is compared with directly fine-tuned DDPMs, modern GAN-based approaches, and few-shot fine-tuning approaches of large text-to-image models on generation quality and diversity qualitatively and quantitatively.

\textbf{Basic Setups} For unconditional image generation, we choose FFHQ \cite{Karras_2020_CVPR} and LSUN Church \cite{yu2015lsun} as source datasets and train DDPMs from scratch on these two datasets for 300K and 250K iterations as source models. As for the target datasets, we employ 10-shot Sketches \cite{wang2008face}, Babies, Sunglasses \cite{Karras_2020_CVPR}, and face paintings by Amedeo Modigliani and Raphael Peale \cite{yaniv2019face} in correspondence to the source domain FFHQ. Besides, 10-shot Haunted houses and Landscape drawings are used as the target datasets in correspondence to LSUN Church. The model setups are consistent with the experiments on small-scale datasets in Sec. \ref{section3}. The adapted models are trained for 3K-5K iterations with a batch size of 24 on $\times 8$ NVIDIA RTX A6000 GPUs. 

For text-to-image generation, we use Stable Diffusion \cite{rombach2021highresolution} v1.4 as the source model. Several few-shot datasets containing no more than 10 images, including Van Gogh houses, Watercolor dogs \cite{sohn2023styledrop}, Inkpaintings, and Watercolor paintings \cite{sohn2023learning} are used as target datasets. The adapted models are trained for 800-1500 iterations with a batch size of 4 on a single NVIDIA RTX A6000 GPU. Our approach shares the same learning rates and iterations with DreamBooth \cite{ruiz2022dreambooth} for fair comparison. We use typical formats like ``a [V] house'' and ``a house in the [V] style'' as text prompts. 


\textbf{Evaluation Metrics} We follow CDC \cite{ojha2021few-shot-gan} to use Intra-LPIPS for generation diversity evaluation. To be more specific, we generate 1000 images and assign them to one of the training samples with the lowest LPIPS \cite{zhang2018unreasonable} distance. Intra-LPIPS is defined as the average pairwise LPIPS distances within members of the same cluster averaged over all the clusters. If a model exactly replicates training samples, its Intra-LPIPS will have a score of zero. Larger Intra-LPIPS values correspond to greater generation diversity. 

FID \cite{heusel2017gans} is widely used to evaluate the generation quality of generative models by computing the distribution distances between generated samples and datasets. However, FID would become unstable and unreliable when it comes to datasets containing a few samples (e.g., 10-shot datasets used in this paper). Therefore, we provide FID evaluation using relatively richer target datasets including Sunglasses and Babies, which contain 2500 and 2700 images for unconditional image generation.

As for text-to-image generation, we depend on CLIP \cite{radford2021learning} to measure the textual alignment with target text prompts and domain/style consistency. We denote two metrics: CLIP-Text \cite{ruiz2022dreambooth} and CLIP-Image. CLIP-Text is the average pairwise cosine similarity between the CLIP embeddings of text prompts and generated samples. CLIP-Image is the average pairwise cosine similarity between the CLIP embeddings of training and generated samples. CLIP-Image may be unbiased when the model is overfitting. If a model exactly replicates training samples, its CLIP-Image will have the highest score of 1. We provide CLIP-Image results as a reference.

We fix noise inputs for DDPM-based and GAN-based approaches respectively to synthesize samples for fair comparison of generation quality and diversity.



\begin{figure*}[t]
    \centering
    \includegraphics[width=1.0\linewidth]{ 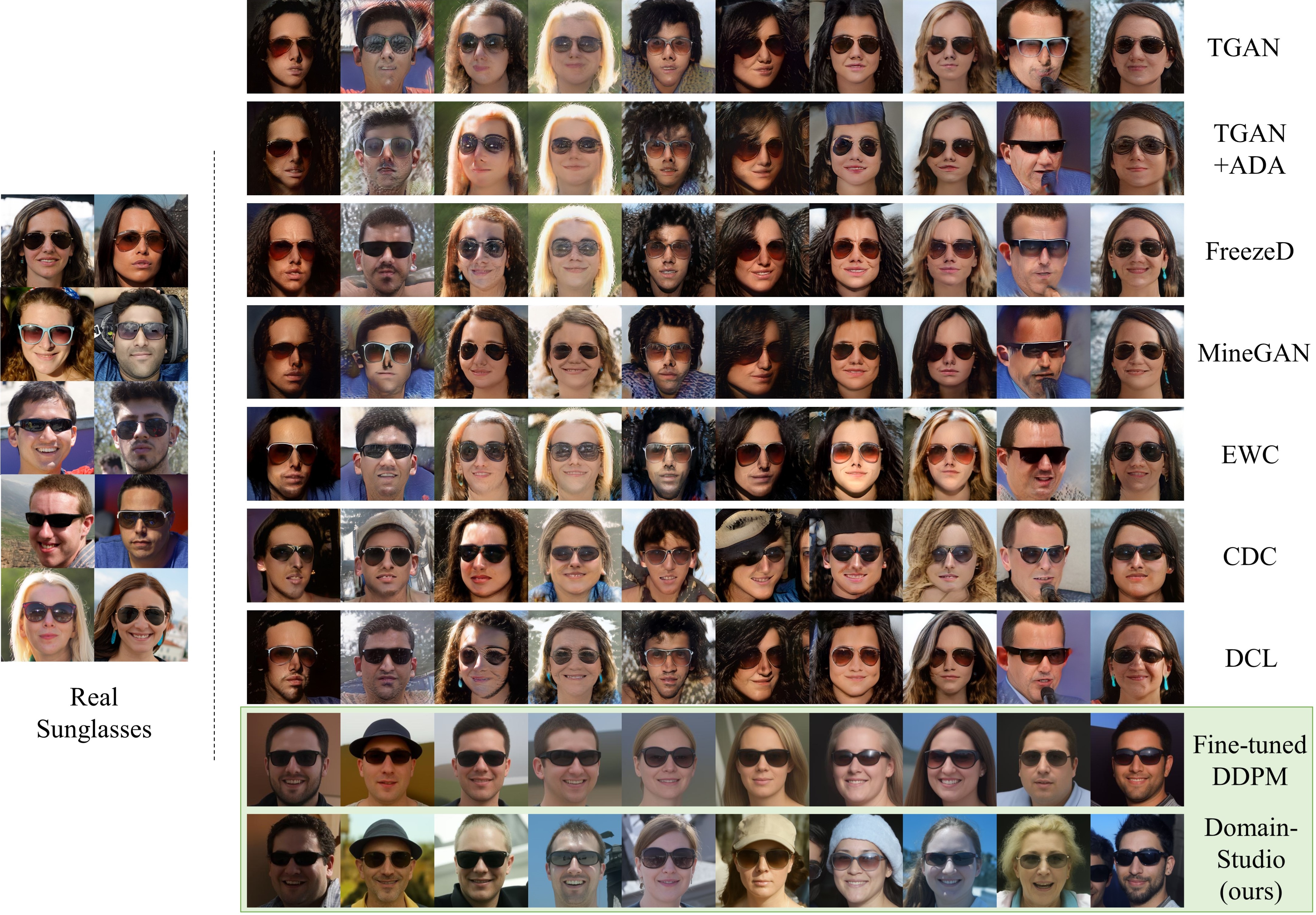}
    \caption{10-shot image generation samples on FFHQ $\rightarrow$ Sunglasses. All the samples of GAN-based approaches are synthesized from fixed noise inputs (rows 1-7). Samples of the directly fine-tuned DDPM and DomainStudio are synthesized from fixed noise inputs as well (rows 8-9). Our approach generates high-quality results with fewer blurs and artifacts and achieves considerable generation diversity.}
    \label{sunglass}
\end{figure*}

\textbf{Baselines}
Since few prior works realize unconditional few-shot image generation with DDPM-based approaches, we employ several GAN-based baselines sharing similar targets with us to adapt pre-trained models to target domains using only a few available samples for comparison: TGAN \cite{wang2018transferring}, TGAN+ADA \cite{ada}, FreezeD \cite{mo2020freeze}, MineGAN \cite{wang2020minegan}, EWC \cite{ewc}, CDC \cite{ojha2021few-shot-gan}, and DCL \cite{zhao2022closer}. All the methods are implemented based on the same StyleGAN2 \cite{Karras_2020_CVPR} codebase. DDPMs directly fine-tuned on limited data are included for comparison as well. The StyleGAN2 models and DDPMs trained on the large source datasets share similar generation quality and diversity (see more details in Appendix \ref{appendix_source}). In addition, modern few-shot fine-tuning methods of text-to-image models including Textual Inversion \cite{gal2022textual} and DreamBooth \cite{ruiz2022dreambooth} are used as baselines for conditional text-to-image generation. Textual Inversion is trained to learn the style of few-shot training samples.

\subsection{Qualitative Evaluation}
\textbf{Unconditional Image Generation}
We visualize the samples of DomainStudio on 10-shot FFHQ $\rightarrow$ Babies, FFHQ $\rightarrow$ Sketches, and LSUN Church $\rightarrow$ Haunted houses in the bottom row of Fig. \ref{result1}. DomainStudio produces more diverse samples containing richer high-frequency details than directly fine-tuned DDPMs. For example, DomainStudio generates babies with various detailed hairstyles and facial features. Fig. \ref{result2} adds visualized samples under other adaptation setups. DomainStudio adapts source models to target domains naturally and preserves various features different from training data. Samples of people wearing hats can be found when adapting FFHQ to Babies or Sunglasses, which is obviously different from the training samples. The adaptation from LSUN Church to Haunted houses and Landscape drawings retain various architectural structures. Fig. \ref{sunglass} shows samples of GAN-based and DDPM-based approaches on 10-shot FFHQ $\rightarrow$ Sunglasses. For intuitive comparison, we fix the noise inputs for GAN-based and DDPM-based approaches, respectively. GAN-based baselines generate samples containing unnatural blurs and artifacts. Besides, many incomplete sunglasses in generated samples lead to vague visual effects. In contrast, the directly fine-tuned DDPM produces smoother results but lacks details. DomainStudio improves generation quality and diversity, achieving more pleasing visual effects than existing approaches. Additional visualized comparison results can be found in Appendix \ref{appendix_results}.%

\begin{figure*}[t]
    \centering
    \includegraphics[width=1.0\linewidth]{ 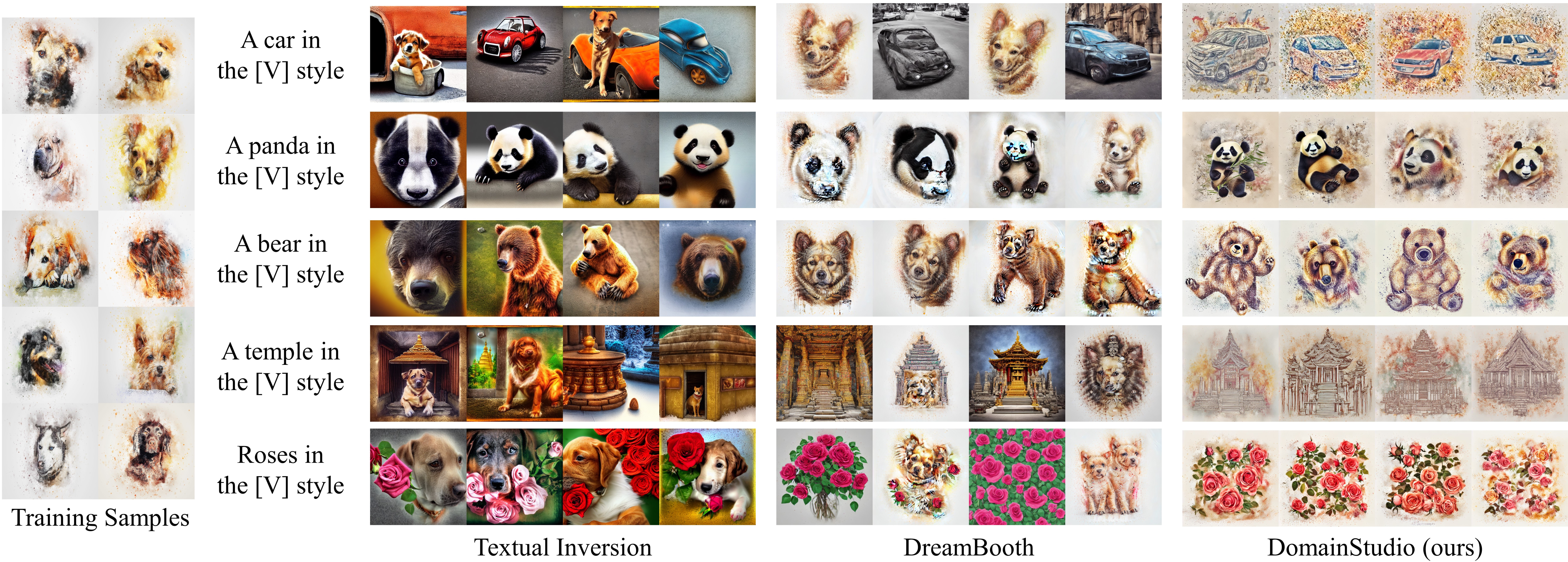}
    \caption{Qualitative comparison of domain-driven text-to-image generation trained on 10-shot Dog watercolors. DomainStudio can produce samples containing diverse subjects consistent with the text prompts and sharing the same style with training samples.}
    \label{stable1}
\end{figure*}

\begin{figure*}[t]
    \centering
    \includegraphics[width=1.0\linewidth]{ 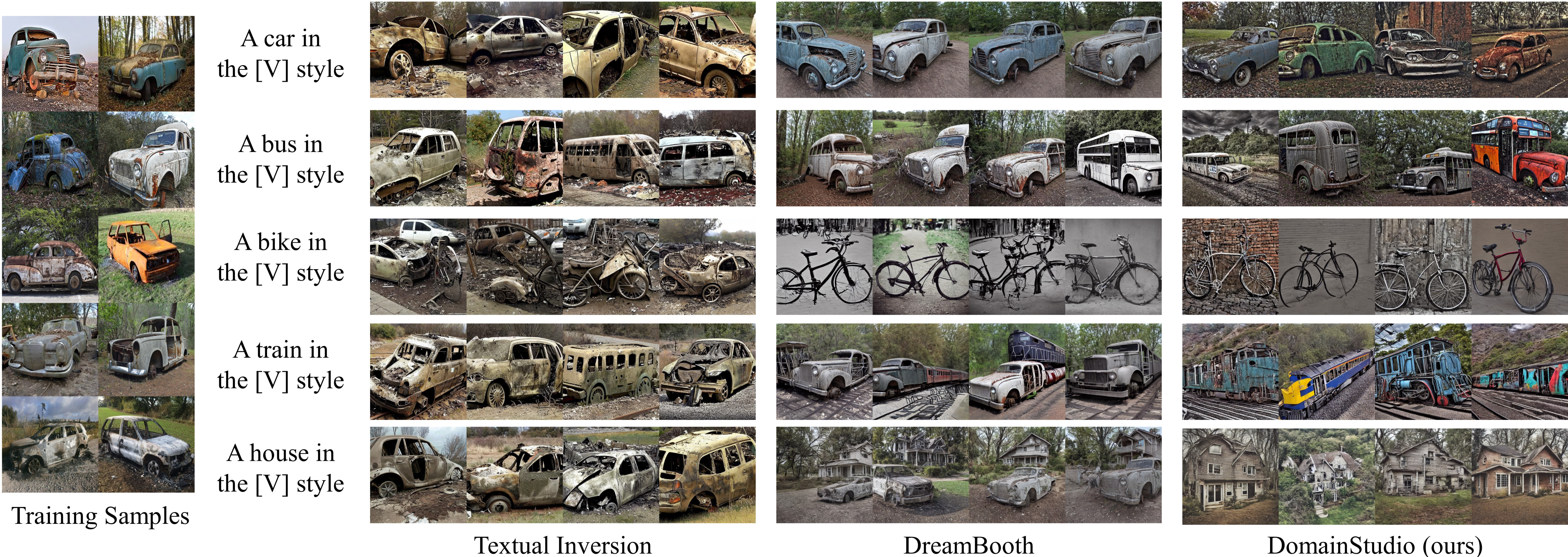}
    \caption{Qualitative comparison of domain-driven text-to-image generation trained on 10-shot Wrecked cars.}
    \label{stable2}
\end{figure*}

\begin{figure*}[t]
    \centering
    \includegraphics[width=1.0\linewidth]{ 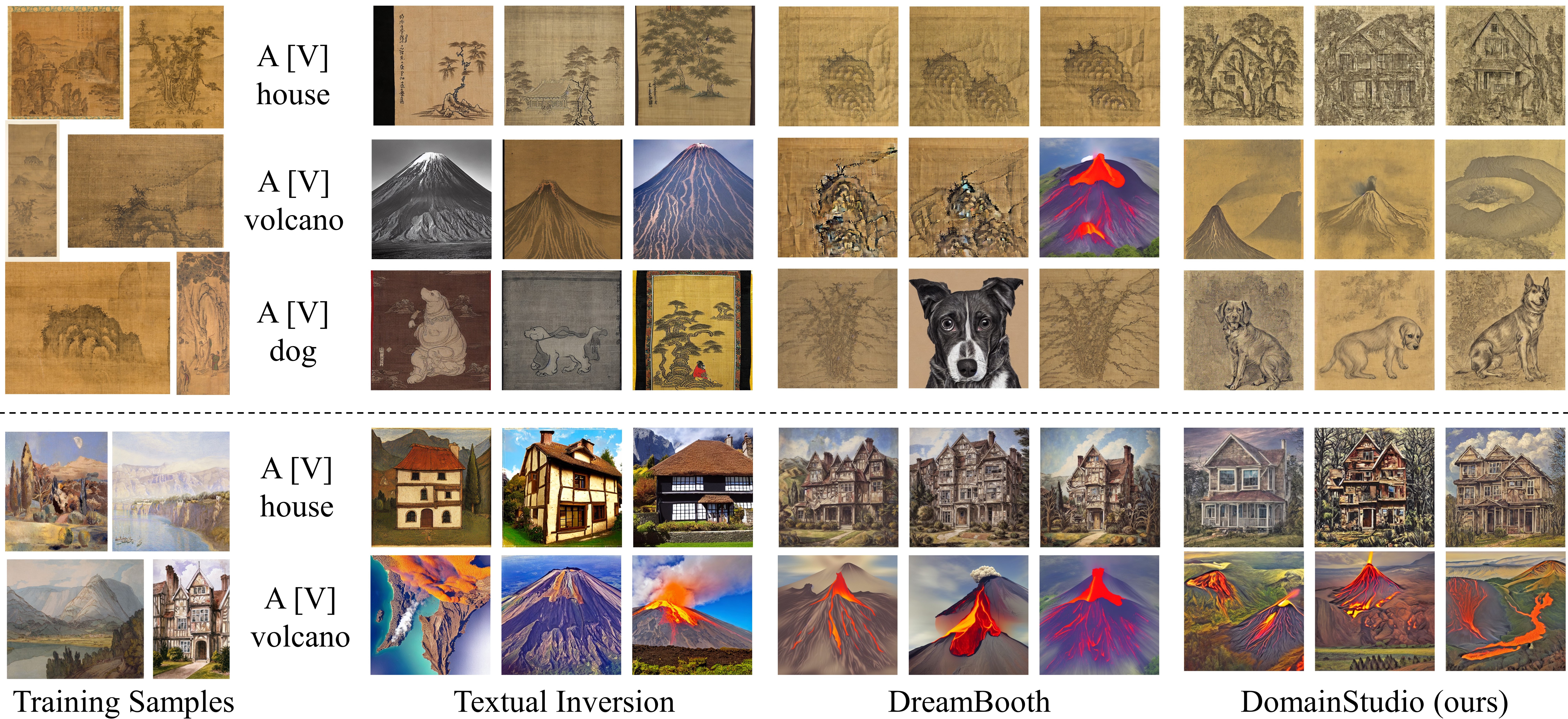}
    \caption{Qualitative comparison of domain-driven text-to-image generation trained on few-shot inkpaintings and watercolors.}
    \label{stable3}
\end{figure*}

\begin{table*}[tbp]
\centering
\setlength{\tabcolsep}{2mm}{
\begin{tabular}{l|c|c|c|c|c}
   Approaches & \makecell[c]{FFHQ $\rightarrow$ \\ Babies} & \makecell[c]{FFHQ $\rightarrow$ \\ Sunglasses} & \makecell[c]{FFHQ $\rightarrow$ \\ Raphael's paintings} & \makecell[c]{LSUN Church $\rightarrow$ \\ Haunted houses} & \makecell[c]{LSUN Church $\rightarrow$ \\ Landscape drawings} 
 \\
\hline
TGAN \cite{wang2018transferring} &  $0.510 \pm 0.026$ & $0.550 \pm 0.021$ & $0.533 \pm 0.023$ & $0.585 \pm 0.007$ & $0.601 \pm 0.030$ \\
TGAN+ADA \cite{ada} & $0.546 \pm 0.033$ & $0.571 \pm 0.034$ & $0.546 \pm 0.037$ &  $0.615 \pm 0.018$ & $0.643 \pm 0.060$ \\
FreezeD \cite{mo2020freeze} &  $0.535 \pm 0.021$ & $0.558 \pm 0.024$ & $0.537 \pm 0.026$ & $0.558 \pm 0.019$ & $0.597 \pm 0.032$ \\
MineGAN \cite{wang2020minegan} & $0.514 \pm 0.034$ & $0.570 \pm 0.020$ & $0.559 \pm 0.031$ & $0.586 \pm 0.041$ & $0.614 \pm 0.027$ \\
EWC \cite{ewc} & $0.560 \pm 0.019$ & $0.550 \pm 0.014$ & $0.541 \pm 0.023$ & $0.579 \pm 0.035$ & $0.596 \pm 0.052$ \\
CDC \cite{ojha2021few-shot-gan} & $0.583 \pm 0.014$ & $0.581 \pm 0.011$ & $0.564 \pm 0.010$ &  $0.620 \pm 0.029$ & $0.674 \pm 0.024$ \\
DCL \cite{zhao2022closer} & $0.579 \pm 0.018$ & $0.574 \pm 0.007$ & $0.558 \pm 0.033$ & $0.616 \pm 0.043$ & $0.626 \pm 0.021$ \\
\hline
Fine-tuned DDPMs & $0.513 \pm 0.026$ & $0.527 \pm 0.024$ & $0.466 \pm 0.018$ & $0.590 \pm 0.045$ &  $0.666 \pm 0.044$ \\
DomainStudio (ours) & $\pmb{0.599 \pm 0.024}$ & $\pmb{0.604 \pm 0.014}$ & $\pmb{0.581 \pm 0.041}$ & $\pmb{0.628 \pm 0.029}$ & $\pmb{0.706 \pm 0.030 }$  \\
\end{tabular}}
\caption{Intra-LPIPS ($\uparrow$) results of DDPM-based approaches and GAN-based baselines on 10-shot image generation tasks adapted from the source domain FFHQ and LSUN Church. Standard deviations are computed across 10 clusters (the same number as training samples). DomainStudio outperforms modern GAN-based approaches and achieves state-of-the-art performance in generation diversity.}
\label{intralpips}
\end{table*}

\textbf{Text-to-image Generation} We visualize the domain-driven text-to-image generation results of DomainStudio using 10-shot Van Gogh houses as training samples in Fig. \ref{vangoghhouse}. Firstly, we use the text prompt ``a [V] house" corresponding to training samples and generate high-quality and diverse target samples. Moreover, we can synthesize samples with different contexts using text prompts like ``a [V] house with night sky". Moreover, we can also synthesize samples in target domains containing different subjects with text prompts such as ``a [V] dog", ``a [V] car" and ``a [V] vase". Finally, we also achieve target samples with diverse contents and contexts different from limited training samples. For example, we synthesize vase samples with various kinds of flowers like sunflowers and roses. Besides, we get vase samples with different contexts following text prompts.

DomainStudio is compared with Textual Inversion and DreamBooth using 10-shot Watercolor dogs and Wrecked cars, 6-shot Inkpaintings, and 4-shot Watercolor paintings in Fig. \ref{stable1}, \ref{stable2}, and \ref{stable3}.  It's difficult for baselines to synthesize reasonable target images with different subjects since they are unable to distinguish what to learn from training samples and what to preserve from source domains. For instance, when generating temples and roses in the watercolor style, baselines tend to combine the subjects in training samples with the subjects mentioned in text prompts, as shown in Fig. \ref{stable1}. DomainStudio guides adapted models to learn the common features of few-shot data and maintain diversity similar to source domains. It successfully adapts the subjects mentioned in text prompts to the style of training samples, even if the subjects are very different from the subjects in training samples. Similar phenomena can be found for baselines trained on wrecked cars, as shown in Fig. \ref{stable2}. Textual Inversion synthesizes car samples with text prompts of train or house. DreamBooth overfits and generates samples similar to few-shot data. It generates train and house samples containing wrecked cars instead of wrecked trains and houses like DomainStudio. Moreover, DomainStudio also achieves compelling results using fewer training samples while baselines either overfit to synthesize low-quality and replicated samples or underfit to synthesize samples inconsistent with target domains, as shown in Fig. \ref{stable3}.


\subsection{Quantitative Evaluation}
\textbf{Unconditional Image Generation}
We provide the Intra-LPIPS results of DomainStudio under a series of 10-shot adaptation setups in Table \ref{intralpips}. DomainStudio realizes a superior improvement of Intra-LPIPS compared with directly fine-tuned DDPMs. Besides, DomainStudio outperforms state-of-the-art GAN-based approaches, indicating its strong capability of maintaining generation diversity. Intra-LPIPS results under other adaptation setups are added in Appendix \ref{appendix_fid}.

\begin{table}[t]
\centering
\setlength\tabcolsep{1.2mm}{
\small
\begin{tabular}{l|c|c|c|c|c|c}
Method & TGAN & ADA & EWC & CDC & DCL & Ours\\
\hline
Babies & $104.79$ & $102.58$ & $87.41$ & $74.39$ & $52.56$ & $ \pmb{33.26}$ \\
Sunglasses & $55.61$ & $53.64$ & $59.73$ & $42.13$ & $38.01$ & $ \pmb{21.92}$\\
\end{tabular}}
\caption{FID ($\downarrow$) results of DomainStudio compared with GAN-based baselines on 10-shot FFHQ $\rightarrow$ Babies and Sunglasses.}
\label{fid}
\end{table}

As shown by the FID results in Table \ref{fid}, DomainStudio performs better on learning target distributions from limited data than prior GAN-based approaches. 

\begin{table*}[tbp]
\centering
\small
\begin{tabular}{l|c|c|c|c|c|c}
Metrics & \multicolumn{3}{c|}{CLIP-Text} & \multicolumn{3}{c}{CLIP-Image}\\ \hline
Datasets & \makecell[c]{Van Gogh \\ houses} & \makecell[c]{Wrecked \\ trains} &\makecell[c]{Inkpainting \\ volcanoes} &  \makecell[c]{Van Gogh \\ houses} & \makecell[c]{Wrecked \\ trains} &\makecell[c]{Inkpainting \\volcanoes} \\
\hline                                                             
 Textual Inversion \cite{gal2022textual} & $0.259 \pm 0.001$ & $0.243 \pm 0.002$ & $0.244 \pm 0.001$ & $0.763 \pm 0.022 $ & $\pmb{0.737 \pm 0.035}$ & $0.658 \pm 0.023$\\
 DreamBooth \cite{ruiz2022dreambooth} & $0.262 \pm 0.002$ & $0.267 \pm 0.001$ & $0.275 \pm 0.004$ & $0.569 \pm 0.039$ & $0.557 \pm 0.011$ & $0.600 \pm 0.086$ \\
 DomainStudio (ours)  & $\pmb{0.276 \pm 0.002}$ & $\pmb{0.271 \pm 0.003}$ & $\pmb{0.301 \pm 0.002}$ & $\pmb{0.789 \pm 0.024}$ & $0.600 \pm 0.068$ & $\pmb{0.676 \pm 0.091}$ \\
\end{tabular}
\caption{CLIP-Text ($\uparrow$) and CLIP-Image results of DomainStudio compared with Textual Inversion and DreamBooth. Our approach outperforms existing methods on text alignment.}
\label{clips_t2i}
\end{table*}
 

\begin{table*}[htbp]
\centering
\small
\setlength{\tabcolsep}{2.5mm}{
\begin{tabular}{l|c|c|c|c|c|c}
   Approaches & \makecell[c]{Van Gogh \\ houses} & \makecell[c]{Watercolor \\ pandas} & \makecell[c]{Watercolor \\ temples} & \makecell[c]{Wrecked \\ cars} & \makecell[c]{Wrecked \\ houses} & \makecell{Inkpainting \\ volcanoes} 
 \\
\hline
Textual Inversion \cite{gal2022textual} &  $0.480 \pm 0.235$ & $\pmb{0.744 \pm 0.031}$ & $\pmb{0.763 \pm 0.033}$ & $0.612 \pm 0.024$ & $0.624 \pm 0.015$ &  $\pmb{0.648 \pm 0.038}$ \\
DreamBooth \cite{ruiz2022dreambooth} & $0.558 \pm 0.009$ & $0.450 \pm 0.099$ & $0.553 \pm 0.082$ &  $0.534 \pm 0.027$ & $0.601 \pm 0.034$ &  $0.535 \pm 0.049$\\
DomainStudio (ours)  &  $\pmb{0.588 \pm 0.012}$ & $0.519 \pm 0.014$ & $0.544 \pm 0.010$ & $\pmb{0.636 \pm 0.012}$ & $\pmb{0.628 \pm 0.017}$ &  $0.633 \pm 0.029$\\
\end{tabular}}
\caption{Intra-LPIPS ($\uparrow$) results of DomainStudio compared with Textual Inversion and DreamBooth.}
\label{intralpips_t2i}
\end{table*}

\textbf{Text-to-image Generation}
We report the CLIP-based metrics of DomainStudio compared with Textual Inversion and DreamBooth in Table \ref{clips_t2i}. DomainStudio achieves better results of CLIP-Text than baselines, indicating its ability to synthesize images consistent with text prompts while adapting to target domains. As for CLIP-Image results, DomainStudio also outperforms baselines on almost all the benchmarks. Textual Inversion achieves the best image alignment on Wrecked trains since it is overfitting to the few-shot training car samples instead of synthesizing train samples.

In Table \ref{intralpips_t2i}, we provide Intra-LPIPS results of DomainStudio and baselines to evaluate the generation diversity. DomainStudio achieves state-of-the-art performance when generating target samples like Van Gogh houses and Wrecked cars. Although Textual Inversion also achieves outstanding generation diversity on target samples like Watercolor pandas and temples, it fails to produce samples sharing the same style with training samples and consistent with text prompts, as shown in Fig. \ref{stable1}.



\begin{figure}[t]
    \centering
    \includegraphics[width=1.0\linewidth]{ 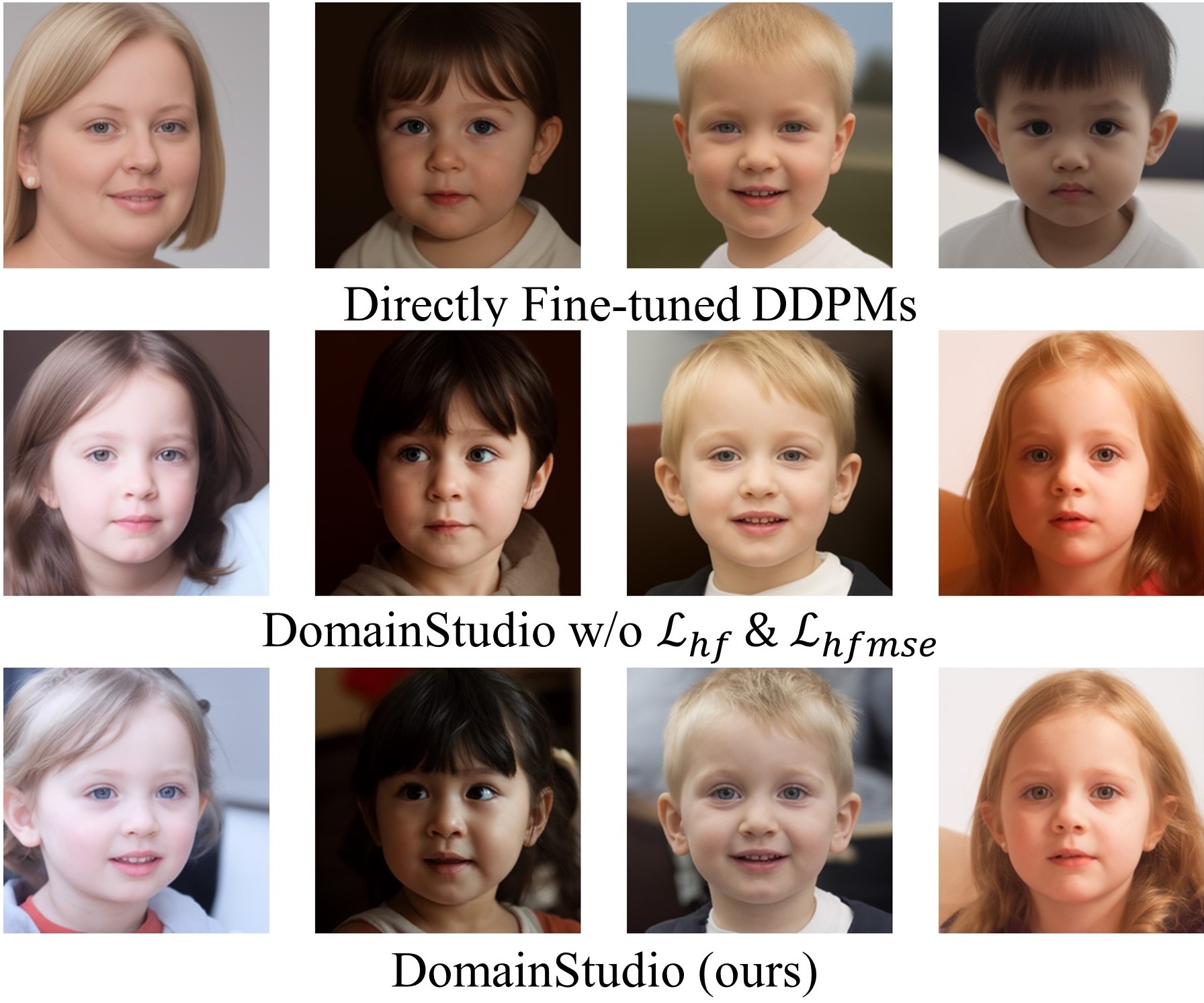}
    \caption{Visualized ablation analysis of the DomainStudio approach using 10-shot FFHQ $\rightarrow$ Babies as an example.}
    \label{ablation}
\end{figure}

\begin{figure}[t]
    \centering
    \includegraphics[width=1.0\linewidth]{ 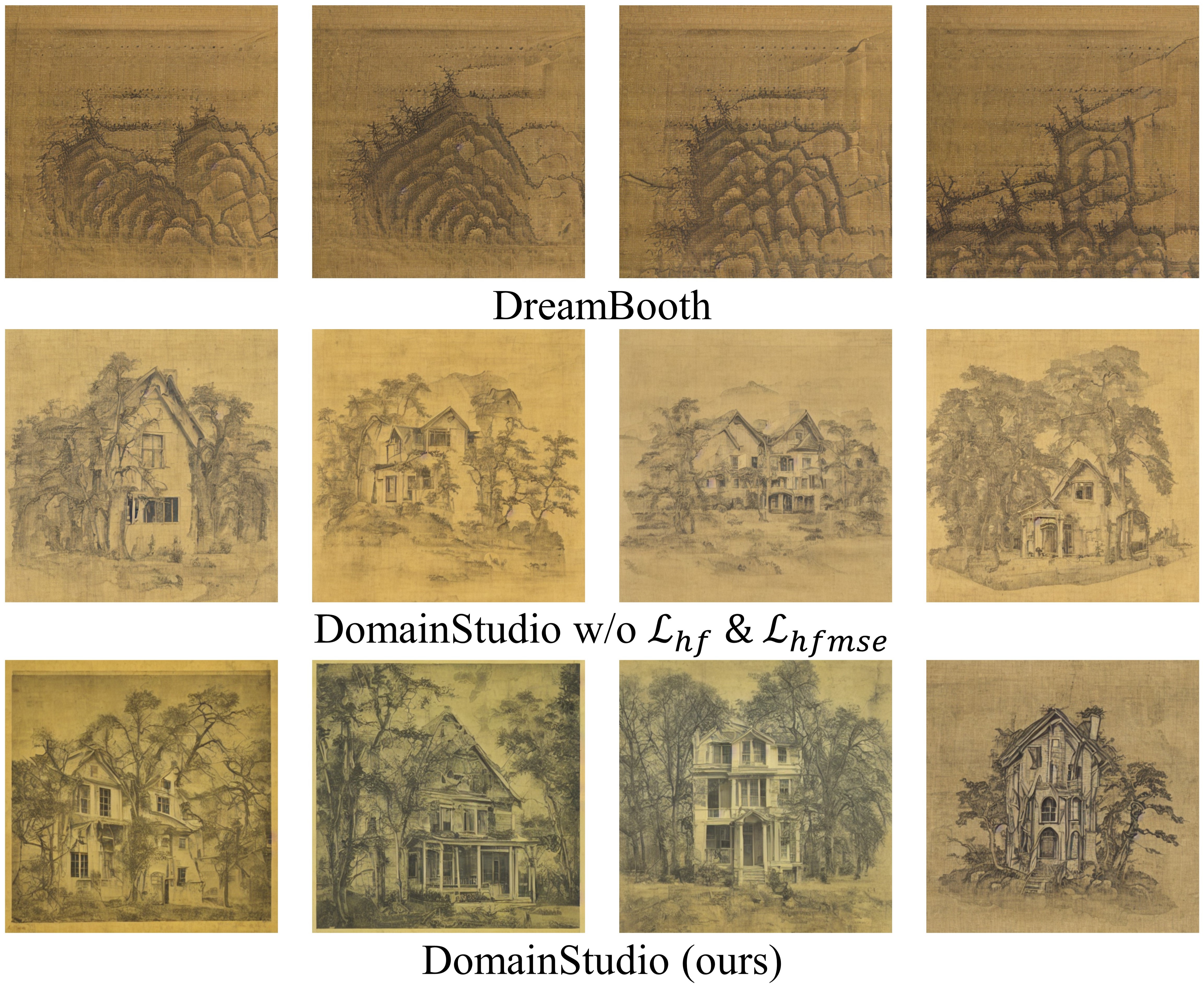}
    \caption{Visualized ablation analysis of the DomainStudio approach using 6-shot Inkpatings as training data and ``a [V] house" as the text prompt.}
    \label{ablation22}
\end{figure}

\subsection{Ablation Analysis}
\label{43}
We ablate our approach on unconditional image generation using 10-shot FFHQ $\rightarrow$ Babies as an example in Fig. \ref{ablation}.  With relative distances preservation only, the adapted model produces coarse samples lacking details like hairstyles. DomainStudio combines both ideas and achieves realistic and diverse results. Ablations of the weight coefficients in Equation \ref{loss} are added in Appendix \ref{appendix_ablation}.

We also provide the visualized ablations of DomainStudio on text-to-image generation using house inkpaintings as an example in Fig. \ref{ablation22}. DreamBooth is designed to preserve key features of training samples. As a result, it overfits and fails to achieve domain-driven generation. DomainStudio without high-frequency details enhancement applies pairwise similarity loss to relieve overfitting and guide adapted models to learn the knowledge of target domains while preserving source subjects corresponding to text prompts. The full DomainStudio approach adds high-frequency details enhancement and preserves more details from source and training samples.

\section{Conclusion}
We propose DomainStudio, a novel approach to realize few-shot and domain-driven image generation with diffusion models. DomainStudio keeps the relative pairwise distances between adapted samples and realizes high-frequency details enhancement during domain adaptation. It is compatible with both unconditional and conditional image generation models. For unconditional image generation, this work first introduces diffusion models to few-shot image generation. We demonstrate the effectiveness of DomainStudio on a series of few-shot image generation tasks. It generates compelling samples with rich details and few blurs, outperforming current state-of-the-art GAN-based approaches on generation quality and diversity. For conditional image generation, this work realizes high-quality domain-driven generation by fine-tuning large text-to-image models with limited data. DomainStudio performs better than existing subject-driven methods and synthesizes samples in target domains with diverse subjects and contexts. We believe that our work takes an essential step toward more data-efficient diffusion models. The limitations are discussed in Appendix \ref{appendix_limitations}. 

\textbf{Societal Impact} DomainStudio proposed in this work could be applied to provide additional data for corner cases needed by downstream tasks and improve the efficiency of artistic creation by synthesizing images containing diverse objects and sharing similar styles with training samples. We recognize that DomainStudio has potential risks of being misused to imitate existing works without permission since it only needs a few samples as training data.

\small
\bibliographystyle{abbrv}
\bibliography{refs}
\normalsize

\clearpage

\appendix


\section{Supplementary Evaluation}
\label{appendix_fid}
In Table \ref{intralpips2}, we provide the additional results of Intra-LPIPS \cite{ojha2021few-shot-gan} on 10-shot FFHQ $\rightarrow$ Sketches and FFHQ $\rightarrow$ Amedeo's paintings (unconditional) as supplements to Table \ref{intralpips}. The proposed DomainStudio approach outperforms modern GAN-based approaches and directly fine-tuned DDPMs on generation diversity under these adaptation setups as well.

FID results of GAN-based baselines in Table \ref{fid} are reported in CDC \cite{ojha2021few-shot-gan} and DCL \cite{zhao2022closer}.

\section{Supplementary Ablation Analysis}
\label{appendix_ablation}
 In this section, we provide the ablation analysis of the weight coefficients of $\mathcal{L}_{img}$, $\mathcal{L}_{hf}$, and $\mathcal{L}_{hfmse}$ using 10-shot FFHQ $\rightarrow$ Babies (unconditional) as an example. Intra-LPIPS and FID are employed for quantitative evaluation. 
 
\begin{figure}[ht]
    \centering
    \includegraphics[width=1.0\linewidth]{ 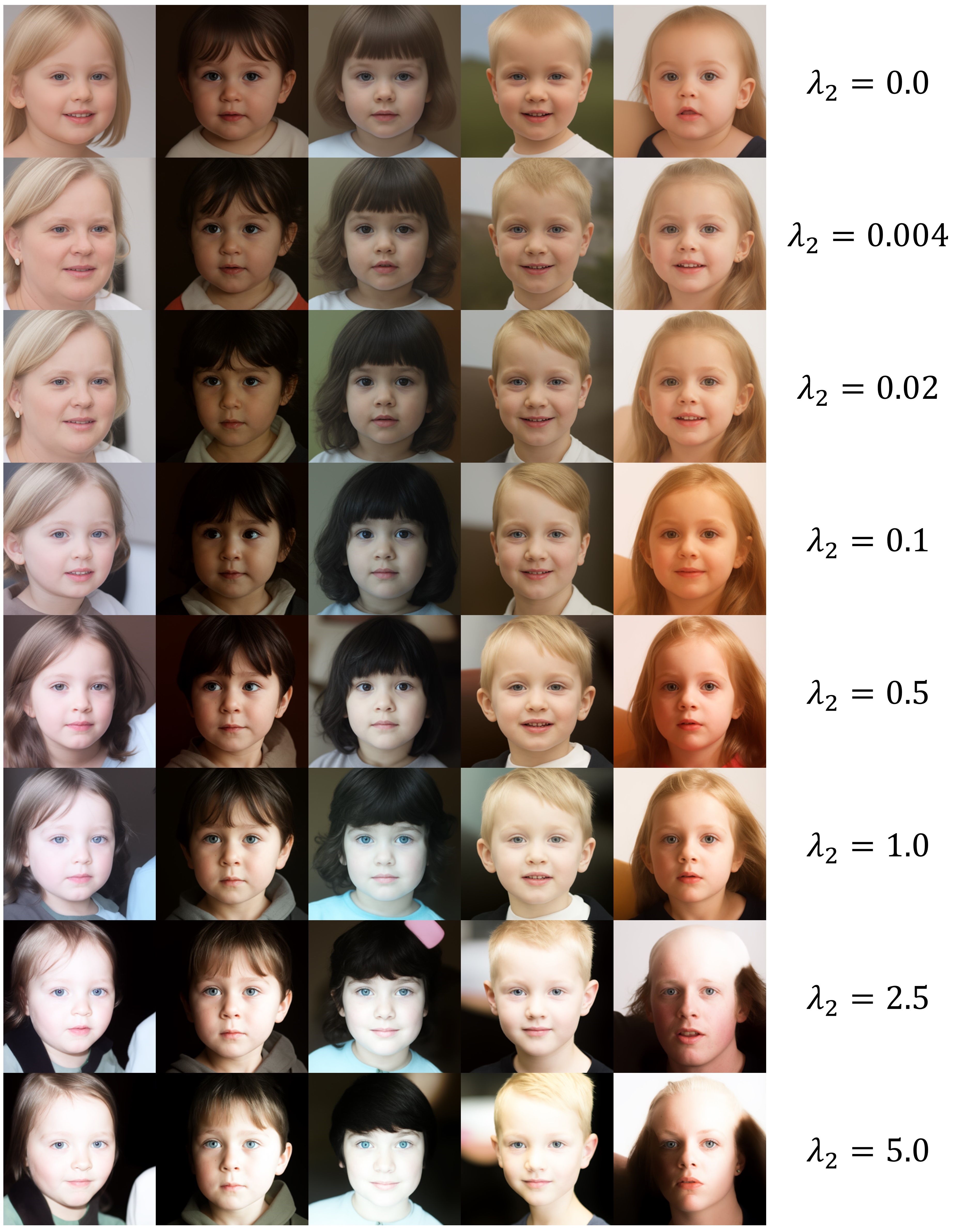}
    \caption{Visualized ablations of $\lambda_2$, the weight coefficient of $\mathcal{L}_{img}$ on 10-shot FFHQ $\rightarrow$ Babies. Samples of different models are synthesized from fixed noise inputs.}
    \label{lambda2_img}
\end{figure}

We first ablate $\lambda_2$, the weight coefficient of $\mathcal{L}_{img}$. We adapt the source model to 10-shot Babies without $\mathcal{L}_{hf}$ and $\mathcal{L}_{hfmse}$. The quantitative results are listed in Table \ref{lambda2}. Corresponding generated samples are shown in Fig. \ref{lambda2_img}. When $\lambda_2$ is set as 0.0, the directly fine-tuned model produces coarse results lacking high-frequency details and diversity. With an appropriate choice of $\lambda_2$, the adapted model achieves greater generation diversity and better learning of target distributions under the guidance of $\mathcal{L}_{img}$. Too large values of $\lambda_2$ make $\mathcal{L}_{img}$ overwhelm $\mathcal{L}_{simple}$ and prevent the adapted model from learning target distributions, leading to degraded generation quality and diversity. The adapted model with $\lambda_2$ value of 2.5 gets unnatural generated samples even if it achieves the best FID result. We recommend $\lambda_2$ ranging from $0.1$ to $1.0$ for the adaptation setups used in our paper based on a comprehensive consideration of the qualitative and quantitative evaluation.

\begin{figure}[ht]
    \centering
    \includegraphics[width=1.0\linewidth]{ 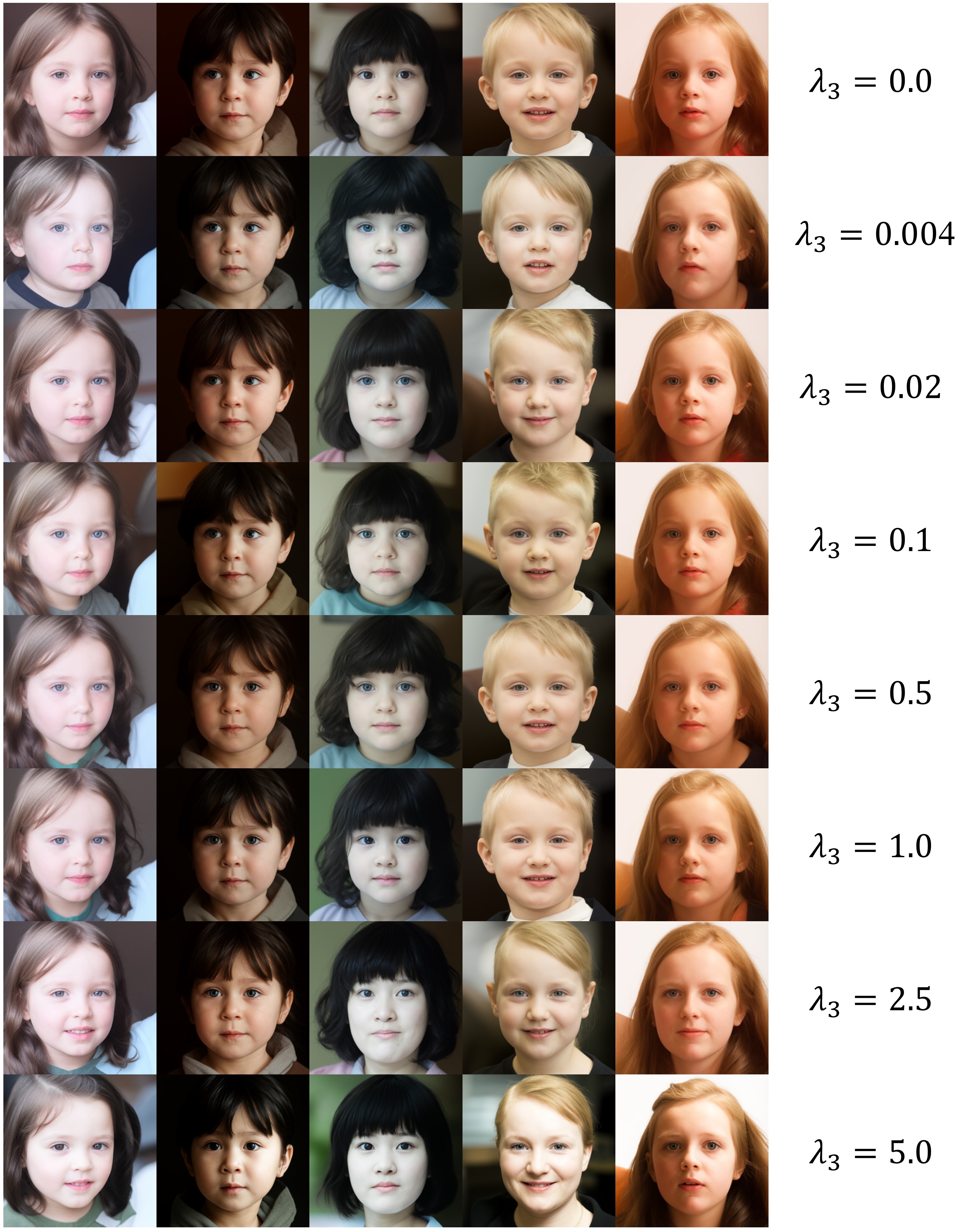}
    \caption{Visualized ablations of $\lambda_3$, the weight coefficient of $\mathcal{L}_{hf}$ on 10-shot FFHQ $\rightarrow$ Babies. Samples of different models are synthesized from fixed noise inputs.}
    \label{lambda3_img}
\end{figure}

Next, we ablate $\lambda_3$, the weight coefficient of $\mathcal{L}_{hf}$ with $\lambda_2$ set as 0.5. The quantitative results are listed in Table \ref{lambda3}. Corresponding generated samples are shown in Fig. \ref{lambda3_img}. $\mathcal{L}_{hf}$ guides adapted models to keep diverse high-frequency details learned from source domains for more realistic results. $\mathcal{L}_{hf}$ helps the adapted model enhance details like clothes and hairstyles and achieves better FID and Intra-LPIPS, indicating improved quality and diversity. Too large values of $\lambda_3$ make the adapted model pay too much attention to high-frequency components and fail to produce realistic results following the target distributions. We recommend $\lambda_3$ ranging from 0.1 to 1.0 for the adaptation setups used in our paper.

\begin{table}[t]
\centering
\setlength\tabcolsep{3.4pt}
\begin{tabular}{l|c|c}
 Approaches & \makecell[c]{FFHQ $\rightarrow$ \\ Sketches} & \makecell[c]{FFHQ $\rightarrow$ \\ Amedeo's paintings} 
 \\
 \hline
 TGAN \cite{wang2018transferring} &  $0.394 \pm 0.023$ & $0.548 \pm 0.026$  \\
 TGAN+ADA \cite{ada} & $0.427 \pm 0.022$ & $0.560 \pm 0.019$  \\
 FreezeD \cite{mo2020freeze} & $0.406 \pm 0.017$ & $0.597 \pm 0.032$  \\
 MineGAN \cite{wang2020minegan} & $0.407 \pm 0.020$ & $0.614 \pm 0.027$  \\
 EWC \cite{ewc} & $0.430 \pm 0.018$ & $0.594 \pm 0.028$  \\
 CDC \cite{ojha2021few-shot-gan} &  $0.454 \pm 0.017$ & $0.620 \pm 0.029$  \\
 DCL \cite{zhao2022closer} & $0.461 \pm 0.021$ & $0.616 \pm 0.043$  \\
 \hline
 Fine-tuned DDPMs  & $0.473 \pm 0.022$ & $0.484 \pm 0.021$ \\
 DomainStudio (ours) &  $\pmb{0.495 \pm 0.024}$ & $\pmb{0.626 \pm 0.022}$ \\
\end{tabular}
\caption{Intra-LPIPS ($\uparrow$) results of DDPM-based approaches and GAN-based baselines on 10-shot FFHQ $\rightarrow$ Sketches and FFHQ $\rightarrow$ Amedeo's paintings. Standard deviations are computed across 10 clusters (the same number as training samples).}
\label{intralpips2}
\end{table}

\begin{table}[tbp]
    \centering
    \begin{tabular}{c|c|c}
        $\lambda_2$ & Intra-LPIPS ($\uparrow$) & FID ($\downarrow$) \\
        \hline
        $0.0$ &  $0.520 \pm 0.026$ & $89.94$ \\
        $0.004$ & $0.531 \pm 0.031$ & $76.17$ \\
        $0.02$ & $0.544 \pm 0.026$ & $69.49$\\
        $0.1$ & $0.558 \pm 0.033$ & $60.00$\\
        $0.5$ & $\pmb{0.572 \pm 0.027}$ & $55.97$\\
        $1.0$ & $0.560 \pm 0.034$ & $58.29$\\
        $2.5$ & $0.543 \pm 0.038$ & $\pmb{46.82}$\\
        $5.0$ & $0.537 \pm 0.028$ & $52.42$\\
    \end{tabular} 
    \caption{Intra-LPIPS ($\uparrow$) and FID ($\downarrow$) results of adapted models trained on 10-shot FFHQ $\rightarrow$ Babies with different $\lambda_2$, the weight coefficient of $\mathcal{L}_{img}$.}
    \label{lambda2}
\end{table}

\begin{table}[tbp]
    \centering
    \begin{tabular}{c|c|c}
        $\lambda_3$ & Intra-LPIPS ($\uparrow$) & FID ($\downarrow$) \\
        \hline
        $0.0$ & $0.572 \pm 0.027$ & $55.97$ \\
        $0.004$ & $0.576 \pm 0.034$ & $\pmb{48.92}$ \\
        $0.02$ & $0.581 \pm 0.045$ & $57.39$ \\
        $0.1$ & $0.589 \pm 0.047$ & $53.14$ \\
        $0.5$ & $\pmb{0.592 \pm 0.031}$ & $53.11$ \\
        $1.0$ & $0.583 \pm 0.032$ & $50.04$ \\
        $2.5$ & $0.577 \pm 0.032 $ & $54.60$ \\
        $5.0$ & $0.591 \pm 0.031 $ & $54.61$ \\
    \end{tabular} 
    \caption{Intra-LPIPS ($\uparrow$) and FID ($\downarrow$) results of adapted models trained on 10-shot FFHQ $\rightarrow$ Babies with different $\lambda_3$, the weight coefficient of $\mathcal{L}_{hf}$.}
    \label{lambda3}
\end{table}

\begin{figure}[ht]
    \centering
    \includegraphics[width=1.0\linewidth]{ 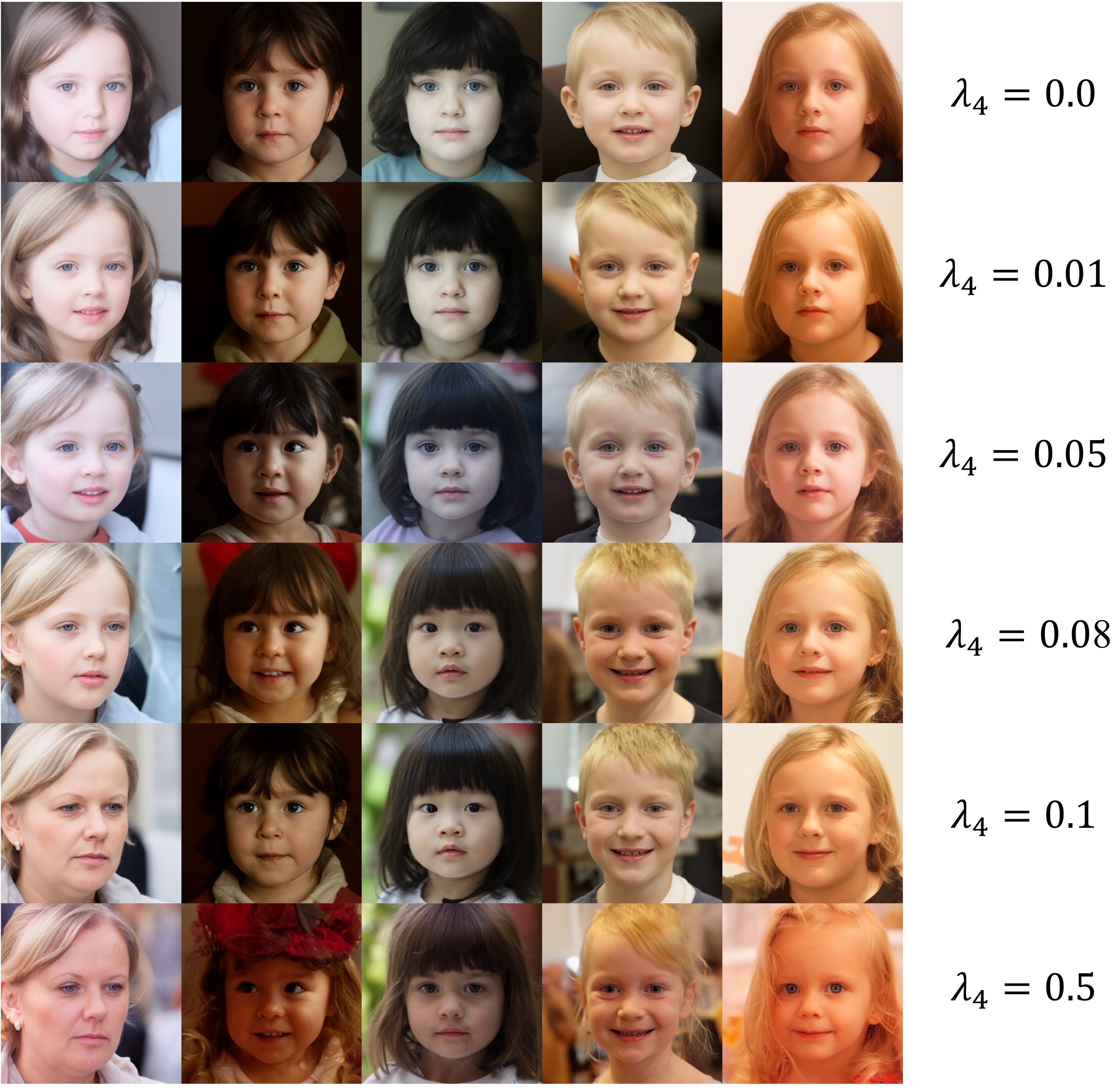}
    \caption{Visualized ablations of $\lambda_4$, the weight coefficient of $\mathcal{L}_{hfmse}$ on 10-shot FFHQ $\rightarrow$ Babies. Samples of different models are synthesized from fixed noise inputs.}
    \label{lambda4_img}
\end{figure}

\begin{table}[tbp]
    \centering
    \begin{tabular}{c|c|c}
        $\lambda_4$ & Intra-LPIPS ($\uparrow$) & FID ($\downarrow$) \\
        \hline
        $0.0$ & $0.592 \pm 0.031$ & $53.11$ \\
        $0.01$ & $0.594 \pm 0.038$ & $47.54$ \\
        $0.05$ & $0.599 \pm 0.024$ & $\pmb{33.26}$\\
        $0.08$ & $0.607 \pm 0.025$ & $ 38.93$\\
        $0.1$ & $0.603 \pm 0.031$ &  $44.23$\\
        $0.5$ & $\pmb{0.612 \pm 0.023}$ & $53.69$\\
    \end{tabular} 
    \caption{Intra-LPIPS ($\uparrow$) and FID ($\downarrow$) results of adapted models trained on 10-shot FFHQ $\rightarrow$ Babies with different $\lambda_4$, the weight coefficient of $\mathcal{L}_{hfmse}$.}
    \label{lambda4}
\end{table}

Finally, we ablate $\lambda_4$, the weight coefficient of $\mathcal{L}_{hfmse}$, with $\lambda_2$ and $\lambda_3$ set as 0.5. The quantitative results are listed in Table \ref{lambda4}. Corresponding generated samples are shown in Fig. \ref{lambda4_img}. $\mathcal{L}_{hfmse}$ guides the adapted model to learn more high-frequency details from limited training data. Appropriate choice of $\lambda_4$ helps the adapted model generate diverse results containing rich details. Besides, the full DomainStudio approach achieves state-of-the-art results of FID and Intra-LPIPS on 10-shot FFHQ $\rightarrow$ Babies (see Table \ref{intralpips} and \ref{fid}). Similar to $\lambda_2$ and $\lambda_3$, too large values of $\lambda_4$ lead to unreasonable results deviating from the target distributions. We recommend $\lambda_4$ ranging from 0.01 to 0.08 for the adaptation setups in this paper.

\begin{figure*}[ht]
    \centering
    \includegraphics[width=1.0\linewidth]{ 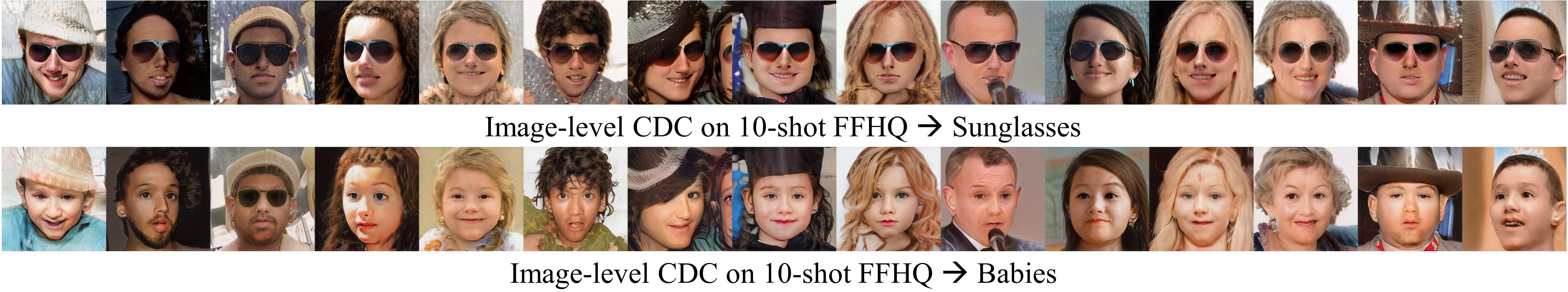}
    \caption{Image samples synthesized by CDC \cite{ojha2021few-shot-gan} using image-level information on 10-shot FFHQ $\rightarrow$ Sunglasses and FFHQ $\rightarrow$ Babies. }
    \label{imagecdc}
\end{figure*}

\section{Limitations}
\label{appendix_limitations}
 Despite the compelling results of our approach, it still has some limitations. All the datasets used in this paper share the resolution of $256\times 256$. The experiments of DomainStudio are conducted on NVIDIA RTX A6000 GPUs (48 GB memory of each). However, the batch size on each GPU is still limited to 3 for unconditional image generation and 4 for text-to-image generation. Therefore, it is challenging to expand our approach to larger image resolution. We will work on more lightweight DDPM-based few-shot image generation approaches. Despite that, the datasets used in this paper have larger resolution than many DDPM-based works \cite{giannone2022few,nichol2021improved,austin2021structured,chen2022analog,kingma2021variational,zhang2022gddim} which use datasets with resolution $32\times 32$ and $64\times 64$. For text-to-image generation, DomainStudio on Stable Diffuison \cite{rombach2021highresolution} can synthesize images with super-resolution (512 $\times$ 512 or 1024 $\times$ 1024).

Although we have designed methods for high-frequency details enhancement and achieved realistic results, there still exists room for further improvement, especially when target domains contain apparently more high-frequency components than source domains (e.g., LSUN Church $\rightarrow$ Haunted houses). Besides, DomainStudio still cannot fully reproduce the styles of some abstract target domains while maintaining generation diversity. Nevertheless, our work first introduces diffusion models to few-shot image generation tasks and has improved generation quality and diversity compared with existing GAN-based approaches. We hope it will be a solid basis for better methods in the future.

\section{Inspiration of Our Approach}
\subsection{$\mathcal{L}_{img} \; \& \; \mathcal{L}_{hf}$}
The proposed pairwise similarity loss designed for DDPMs is mainly inspired by the methods in contrastive learning \cite{oord2018representation, he2020momentum, chen2020simple}, which build probability distributions based on similarities. Similar methods are applied to GAN-based approaches, including CDC \cite{ojha2021few-shot-gan} and DCL \cite{zhao2022closer} as well. The pairwise similarity loss maintains generation diversity by keeping the distributions of relative pairwise distances between adapted samples similar to source samples during domain adaptation. In this way, the generated samples are prevented from being too similar to each other or replicating training samples. Instead, the adapted model is encouraged to learn the common features of training samples and preserve information learned from source domains, improving generation quality and diversity. 

GAN-based approaches depend on perceptual features in the generator and discriminator to compute similarity and probability distributions. As for the proposed DomainStudio approach, the predicted input images $\tilde{x}_0$ calculated in terms of $x_t$ and $\epsilon_{\theta}(x_t,t)$ (Equation \ref{eq11}) are applied in replacement of perceptual features used for GANs. Besides, the high-frequency components of $\tilde{x}_0$ are applied to pairwise similarity loss calculation for high-frequency details enhancement. DomainStudio directly uses image-level information to preserve the relative pairwise distances between adapted samples and during domain adaptation. 

We tried to use features in diffusion processes (Design A) and images of several diffusion steps (Design B) for pairwise similarity loss calculation. As shown in Table \ref{compar} (FID evaluation on FFHQ $\rightarrow$ Sunglasses, Intra-LPIPS evaluation on 10-shot FFHQ $\rightarrow$ Sunglasses), the proposed loss design using image-level information directly is simple, effective, inexpensive, and achieves the best quality and diversity. Here we do not include high-frequency details enhancement for fair comparison.

\begin{table}[t]
\centering
\small
\begin{tabular}{l|c|c|c}
    Method & FID ($\downarrow$) & Intra-LPIPS ($\uparrow$) & Time $/\ $ 1K iters ($\downarrow$) \\
    \hline
    Ours & $\pmb{37.92}$ & $\pmb{0.59 \pm 0.02}$ & $\pmb{34 min}$ \\ 
    Design A & $40.30$ & $0.55 \pm 0.03$ & $52 min$\\ 
    Design B & $58.28$ & $0.57 \pm 0.06$ & $ 38 min$ \\
\end{tabular} 
\caption{Quantitative evaluation comparison between different designs for the pairwise similarity loss.}
\label{compar}
\end{table}


As illustrated in Sec. \ref{422}, DomainStudio synthesizes more realistic images with fewer blurs and artifacts and achieves better generation diversity than current state-of-the-art GAN-based approaches \cite{ojha2021few-shot-gan, zhao2022closer}. We also try to use image-level information to replace the perceptual features for the GAN-based approach CDC \cite{ojha2021few-shot-gan}. However, we fail to avoid generating artifacts or achieve higher generation quality, as shown in Fig. \ref{imagecdc}. The proposed pairwise similarity loss matches better with diffusion models than GANs. 

\begin{figure*}[ht]
    \centering
    \includegraphics[width=1.0\linewidth]{ 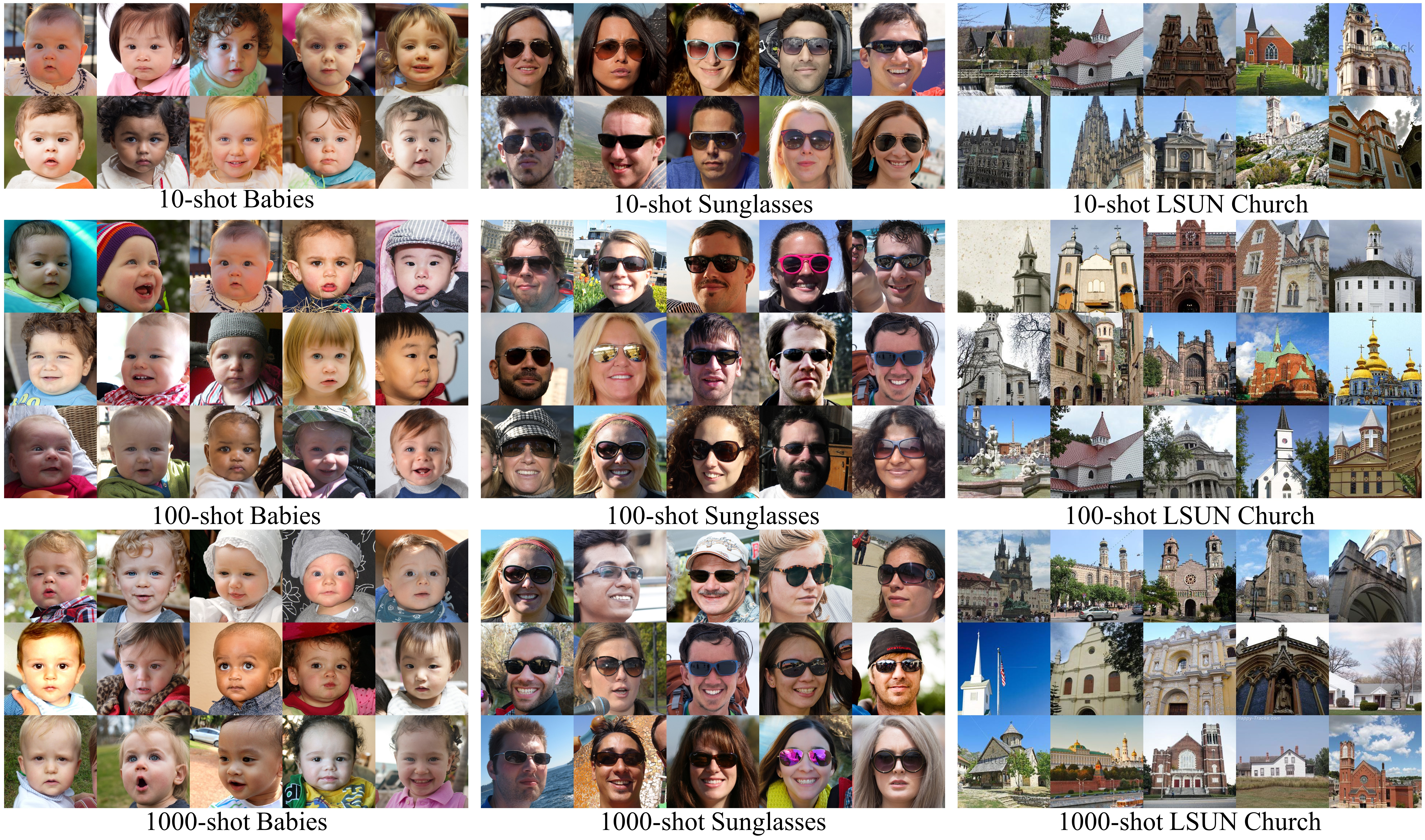}
    \caption{Examples of small-scale datasets sampled from Babies, Sunglasses, and LSUN Church. For datasets containing 100 or 1000 images, we randomly pick 15 examples.}
    \label{scratch_data}
\end{figure*}

\subsection{$\mathcal{L}_{hfmse}$}
DDPMs learn target distributions mainly through mean values of predicted noises using the reweighted loss function (Equation \ref{loss_simple}). As a result, it is hard for DDPMs to learn high-frequency distributions from limited data, as shown in the smooth samples produced by models trained on limited data from scratch in Fig. \ref{scratch2}. Therefore, we propose $\mathcal{L}_{hfmse}$ to strengthen the learning of high-frequency details from limited data during domain adaptation.

\subsection{Full Approach}
Prior GAN-based approaches like CDC \cite{ojha2021few-shot-gan} and DCL \cite{zhao2022closer} aim to build a one-to-one correspondence between source and target samples. DomainStudio focuses on generating realistic and diverse results following target distributions. Building a one-to-one correspondence is not the first consideration of DomainStudio.

\section{Comparison with ZADIS and StyleDrop}
ZADIS \cite{sohn2023learning} and StyleDrop \cite{sohn2023styledrop} are contemporary to this paper and share similar targets with the proposed DomainStudio approach. ZADIS is based on MaskGIT \cite{chang2022maskgit} and learns visual prompts for target domains/styles. In this way, ZADIS realizes compositional image synthesis with disentangled prompts for style and subjects. StyleDrop is based on MUSE \cite{chang2023muse} and synthesizes images with user-provided styles using reference images and descriptive style descriptors for training under the guidance of CLIP \cite{radford2021learning} scores and human feedback. DomainStudio is designed for diffusion models and compatible with typical unconditional DDPMs \cite{sohl2015deep,NEURIPS2020_4c5bcfec} and modern large text-to-image models like Stable Diffusion \cite{rombach2021highresolution}. DomainStudio aims to learn the common features of target domains, which may be artistic styles or properties like sunglasses.

\begin{figure*}[tbp]
    \centering
    \includegraphics[width=1.0\linewidth]{ 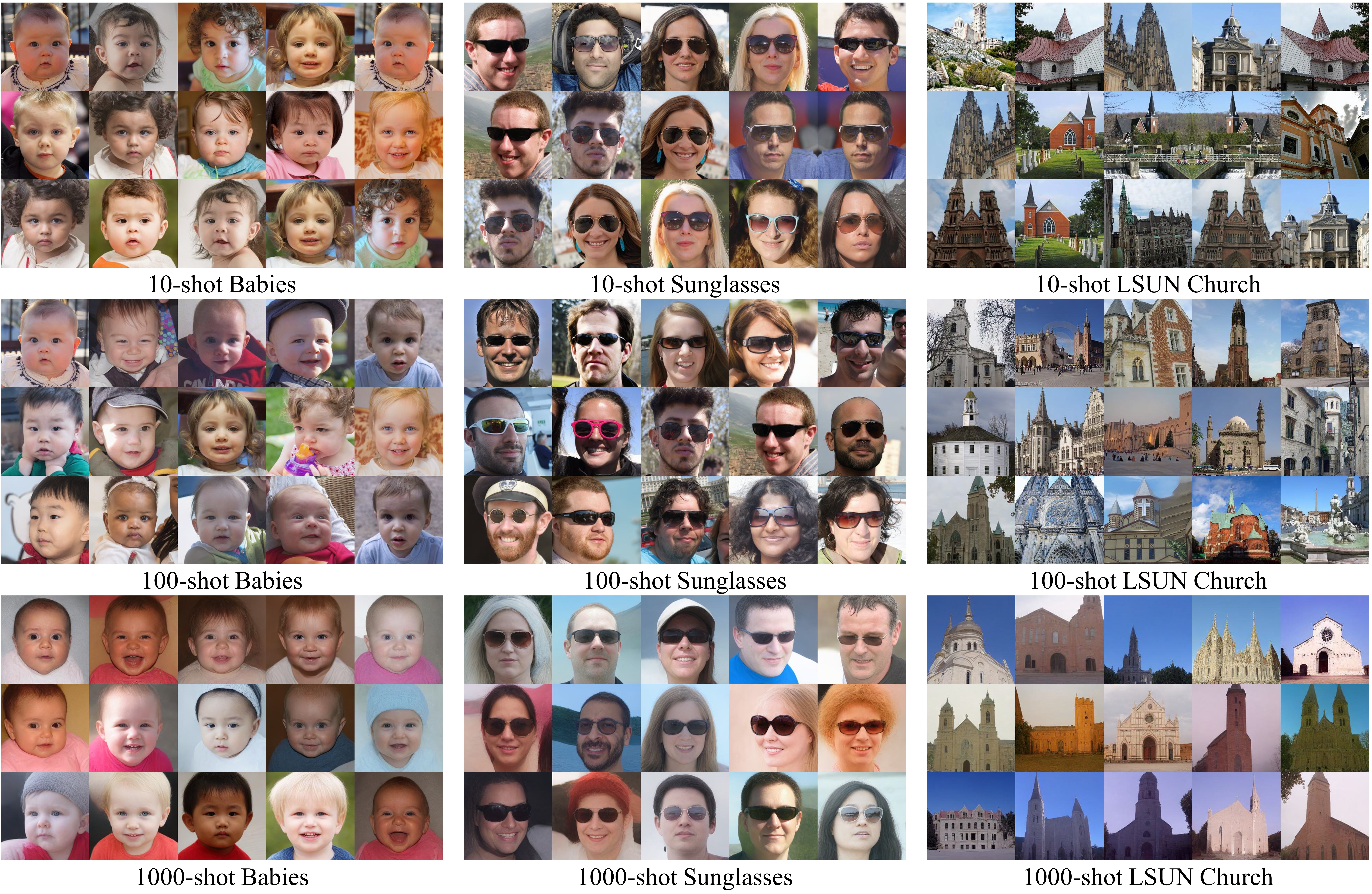}
    \caption{Image samples produced by DDPMs trained from scratch on small-scale datasets, including Babies, Sunglasses, and LSUN Church containing 10, 100, and 1000 images. }
    \label{scratch2}
\end{figure*}

\section{More Details of Implementation}
\label{appendix_implementation}
\subsection{Unconditional DDPMs} 
We follow the model setups of DDPMs used in prior works \cite{nichol2021improved} for LSUN $256^2$ \cite{yu2015lsun} datasets. All the DDPM-based models used in this paper are implemented based on the same codebase \cite{nichol2021improved,dhariwal2021diffusion} and share the same model structure for fair comparison under different adaptation setups and optimization targets. All the source and target datasets are modified to the resolution of $256 \times 256$. We use a max diffusion step $T$ of 1000 and a dropout rate of 0.1. The models are trained to learn the variance with $\mathcal{L}_{vlb}$. The Adam optimizer \cite{kingma2014adam} is employed to update the trainable parameters. We set the learning rate as 0.001 and apply the linear noise addition schedule. Besides, we use half-precision (FP16) binary floating-point format to save memory and make it possible to use a larger batch size in our experiments (batch size 6 for directly fine-tuned DDPMs and batch size 3 for DomainStudio per NVIDIA RTX A6000 GPU). All the results produced by DDPM-based models in this paper follow the sampling process proposed in Ho et al.'s work \cite{NEURIPS2020_4c5bcfec} (about 21 hours needed to generate 1000 samples on a single NVIDIA RTX A6000 GPU) without any fast sampling methods \cite{song2020denoising, zhang2022gddim, lu2022dpm, lu2022dpm2, zhang2022fast, karras2022elucidating}. The weight coefficient $\lambda_2$, $\lambda_3$, and $\lambda_4$ are set as 0.5, 0.5, 0.05 for the quantitative evaluation results of DomainStudio listed in Table \ref{intralpips}, \ref{fid}, and \ref{intralpips2}.

\subsection{Text-to-Image DDPMs}
We follow DreamBooth \cite{ruiz2022dreambooth} to set the learning rates of DomainStudio ranging from 1e-6 to 5e-6. Experiments of DreamBooth and DomainStudio share the same hyperparameters in training for fair comparison. Textual Inversion \cite{gal2022textual} sets the learning rate as 5e-4 and trains text prompts for 2K-3K iterations. The image resolution used for training is 256$\times$256.

\subsection{GAN-based Baselines}
We employ several GAN-based few-shot image generation approaches as baselines for comparison with the proposed DomainStudio approach. Here we provide more details of these baselines. We implement all these approaches based on the same codebase of StyleGAN2 \cite{Karras_2020_CVPR}. The source models are fine-tuned directly on the target datasets to realize TGAN \cite{wang2018transferring}. TGAN+ADA applies ADA \cite{ada} augmentation method to the TGAN baseline. For FreezeD \cite{mo2020freeze}, the first 4 high-resolution layers of the discriminator are frozen following the ablation analysis provided in their work. The results of MineGAN \cite{wang2020minegan} and CDC \cite{ojha2021few-shot-gan} are produced through their official implementation. As for EWC \cite{ewc} and DCL \cite{zhao2022closer}, we implement these approaches following formulas and parameters in their papers since there is no official implementation. These GAN-based approaches are designed for generators \cite{wang2020minegan,ewc,ojha2021few-shot-gan,zhao2022closer} and discriminators \cite{ada,mo2020freeze,zhao2022closer} specially and cannot be expanded to DDPMs directly.

  \begin{figure}[t]
    \centering
    \includegraphics[width=1.0\linewidth]{ 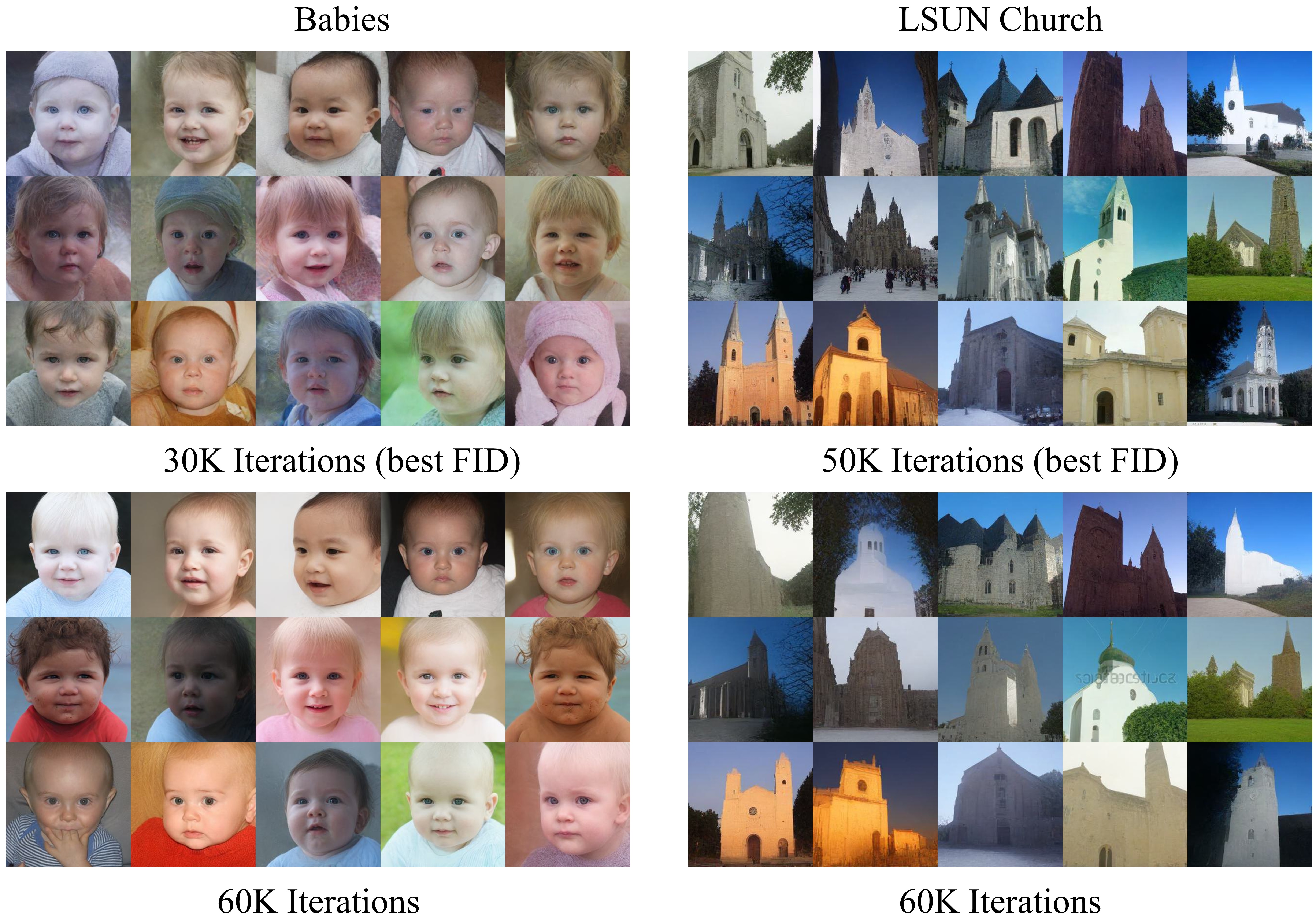}
    \caption{Visualized comparison between DDPMs that achieve the best FID results and DDPMs trained for 60K iterations on 1000-shot Babies and LSUN Church. Image samples produced by different models are synthesized from fixed noise inputs.}
    \label{scratch_compare}
\end{figure}

\begin{figure*}[tbp]
    \centering
    \includegraphics[width=1.0\linewidth]{ 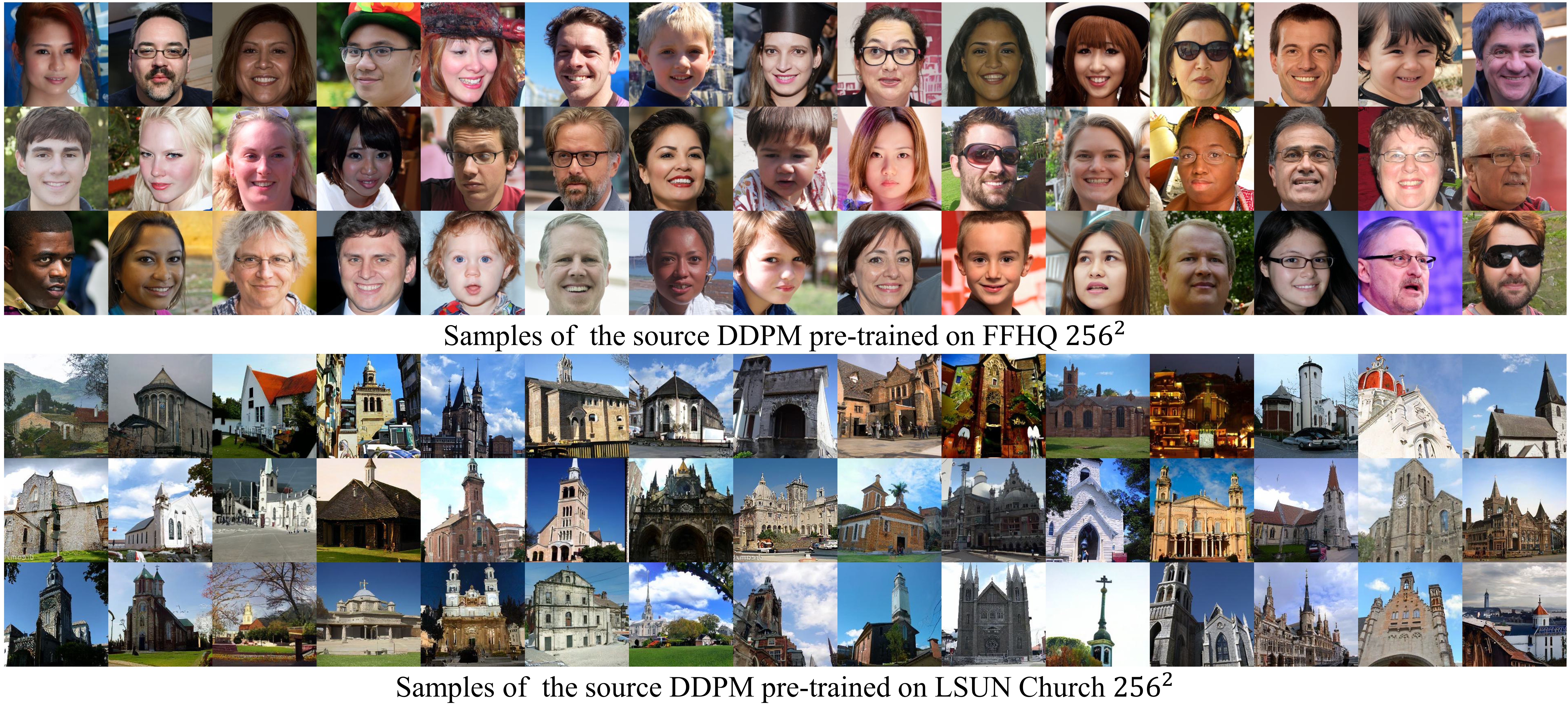}
    \caption{Image samples produced by DDPMs trained on FFHQ $256^2$ \cite{Karras_2020_CVPR} (300K iterations) and LSUN Church $256^2$ \cite{yu2015lsun} (250K iterations).}
    \label{result_scratch}
\end{figure*}

\begin{figure*}[tbp]
    \centering
    \includegraphics[width=1.0\linewidth]{ 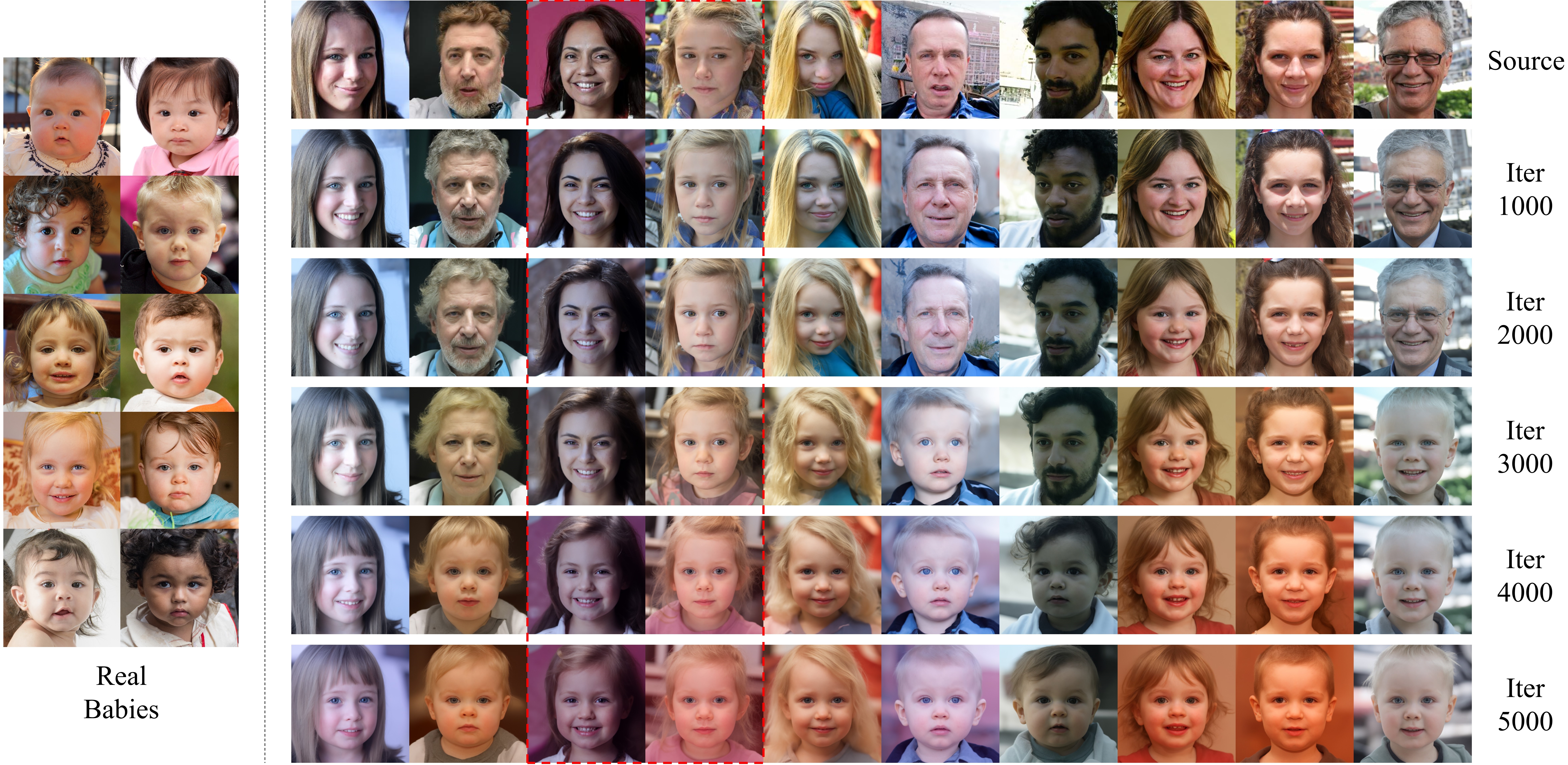}
    \caption{Image samples produced by DomainStudio models trained for different iterations on 10-shot FFHQ $\rightarrow$ Babies. All the visualized samples of different models are synthesized from fixed noise inputs. }
    \label{amedeo}
\end{figure*}

The adapted GANs are trained for 1K-3K iterations. When evaluating generation diversity using Intra-LPIPS \cite{ojha2021few-shot-gan} for the 10-shot adaptation tasks listed in Tables \ref{intralpips} and \ref{intralpips2}, we apply fixed noise inputs to different GAN-based approaches for fair comparison. 

Apart from the GAN-based baselines illustrated above, RSSA \cite{xiao2022few} and AdAM \cite{adaptative} provide different research perspectives for few-shot image generation. RSSA \cite{xiao2022few} preserves the image structure learned from source domains with a relaxed spatial structural alignment method, which is inappropriate for abstract target domains like artists' paintings. AdAM \cite{adaptative} focuses on unrelated source/target domains with an adaptation-aware kernel modulation approach.

\section{DDPMs Trained from Scratch}
\label{appendix_scratch}
In Section \ref{section3}, we evaluate the performance of DDPMs trained from scratch on small-scale datasets containing 10, 100, and 1000 images. In our experiments, the smaller datasets are included in the larger datasets. For example, 1000-shot Sunglasses includes all the images in 100-shot and 10-shot Sunglasses. Similarly, all the images in 10-shot Sunglasses are included in 100-shot Sunglasses as well. We train DDPMs for 40K iterations (about 20 hours on $\times 8$ NVIDIA RTX A6000 GPUs) on datasets containing 10 or 100 images. While for datasets containing 1000 images, DDPMs are trained for 60K iterations (about 30 hours on $\times 8$ NVIDIA RTX A6000 GPUs). 

\begin{table}[tbp]
    \centering
    \begin{tabular}{c|c|c|c}
        Iterations (K) & Babies & Sunglasses & \makecell[c]{LSUN Church}  \\
        \hline
        $0$ & $444.35$ & $419.75$ & $424.53$ \\
        $10$ & $444.91$ & $419.38$ & $413.68$ \\
        $20$ & $222.81$ & $348.48$ & $385.25$ \\
        $30$ & $\pmb{90.16}$ & $168.62$ & $388.81$ \\
        $40$ & $124.97$ & $82.48$ & $57.68$ \\
        $50$ & $132.33$ & $68.26$ & $\pmb{45.43}$ \\
        $60$ & $132.32$ & $\pmb{66.09}$ & $69.18$ \\
    \end{tabular} 
    \caption{FID ($\downarrow$) results of DDPMs trained for different iterations from scratch on 1000-shot Babies, Sunglasses, and LSUN Church.}
    \label{fid_table}
\end{table}

 We provide several typical examples randomly picked from the small-scale datasets in Fig. \ref{scratch_data}. Compared with the generated images shown in Fig. \ref{scratch2}, it can be seen that DDPMs trained from scratch need enough training samples (e.g., 1000 images) to synthesize diverse results and avoid replicating the training samples. Detailed FID \cite{heusel2017gans} results of DDPMs trained from scratch on Babies, Sunglasses, and LSUN Church containing 1000 images for 60K iterations are shown in Table \ref{fid_table}. DDPMs trained from scratch on 1000-shot Babies, Sunglasses, and LSUN Church achieve the best FID results at 30K, 60K, and 50K iterations, respectively.

 In Fig. \ref{scratch2}, we provide samples generated by DDPMs trained from scratch for 60K iterations on all three 1000-shot Babies and LSUN Church datasets. In addition, we add the generated samples of models trained from scratch on 1000-shot Babies and LSUN Church, which achieve the best FID results for comparison in Fig. \ref{scratch_compare}. We do not include samples for all three datasets since the model trained for 60K iterations on 1000-shot Sunglasses achieves the best FID result as well. The model trained on 1000-shot Babies for 60K iterations achieves smoother results containing fewer blurs despite its worse FID result, while the model trained for 30K iterations achieves the best FID result and synthesizes more diverse images. As for the 1000-shot LSUN Church, the model trained for 50K iterations with the best FID result produces samples containing more detailed structures of churches than the model trained for 60K iterations. However, all these results are still coarse and lack high-frequency details, indicating the necessity of adapting source models to target domains when training data is limited.


\begin{table}[tbp]
    \centering
    \begin{tabular}{l|c|c}
        Approaches & FFHQ & LSUN Church  \\
        \hline
        StyleGAN2 & $0.6619 \pm 0.0581$   &  $ 0.7144 \pm 0.0537$ \\
        DDPM &   $\pmb{0.6631 \pm 0.0592}$   & $\pmb{0.7153 \pm 0.0513}$
    \end{tabular} 
    \caption{Average pairwise LPIPS ($\uparrow$) results of 1000 samples produced by StyleGAN2 and DDPMs trained on FFHQ $256^2$ and LSUN Church $256^2$.}
    \label{source_lpips}
\end{table}

\begin{table}[tbp]
    \centering
    \begin{tabular}{l|c|c}
        Approaches & FFHQ & LSUN Church  \\
        \hline
        StyleGAN2 &  $7.71$  & $8.09$   \\
        DDPM &  $\pmb{7.00}$  & $\pmb{6.06}$
    \end{tabular} 
    \caption{FID ($\downarrow$) results of StyleGAN2 and DDPMs trained on FFHQ $256^2$ and LSUN Church $256^2$.}
    \label{source_fid}
\end{table}

\section{DDPM-based Source Models}
\label{appendix_source}
We train DDPMs on FFHQ $256^2$ \cite{Karras_2020_CVPR} and LSUN Church $256^2$ \cite{yu2015lsun} from scratch for 300K iterations and 250K iterations as source models for DDPM adaptation, which cost 5 days and 22 hours, 4 days and 22 hours on $\times 8$ NVIDIA RTX A6000 GPUs, respectively.

Image samples produced by these two source models can be found in Fig. \ref{result_scratch}. We randomly sample 1000 images with these two models to evaluate their generation diversity using the average pairwise LPIPS \cite{zhang2018unreasonable} metric, as shown in Table \ref{source_lpips}. For comparison, we also evaluate the generation diversity of the source StyleGAN2 \cite{Karras_2020_CVPR} models used by GAN-based baselines \cite{wang2018transferring, ada, mo2020freeze, wang2020minegan, ewc, ojha2021few-shot-gan, zhao2022closer}. DDPMs trained on FFHQ $256^2$ and LSUN Church $256^2$ achieve generation diversity similar to the widely-used StyleGAN2 models.

Besides, we sample 5000 images to evaluate the generation quality of the source models using FID \cite{heusel2017gans}. As shown in Table \ref{source_fid}, DDPM-based source models achieve FID results similar to StyleGAN2 on the source datasets FFHQ $256^2$ and LSUN Church $256^2$.

 \begin{figure}[t]
    \centering
    \includegraphics[width=1.0\linewidth]{ 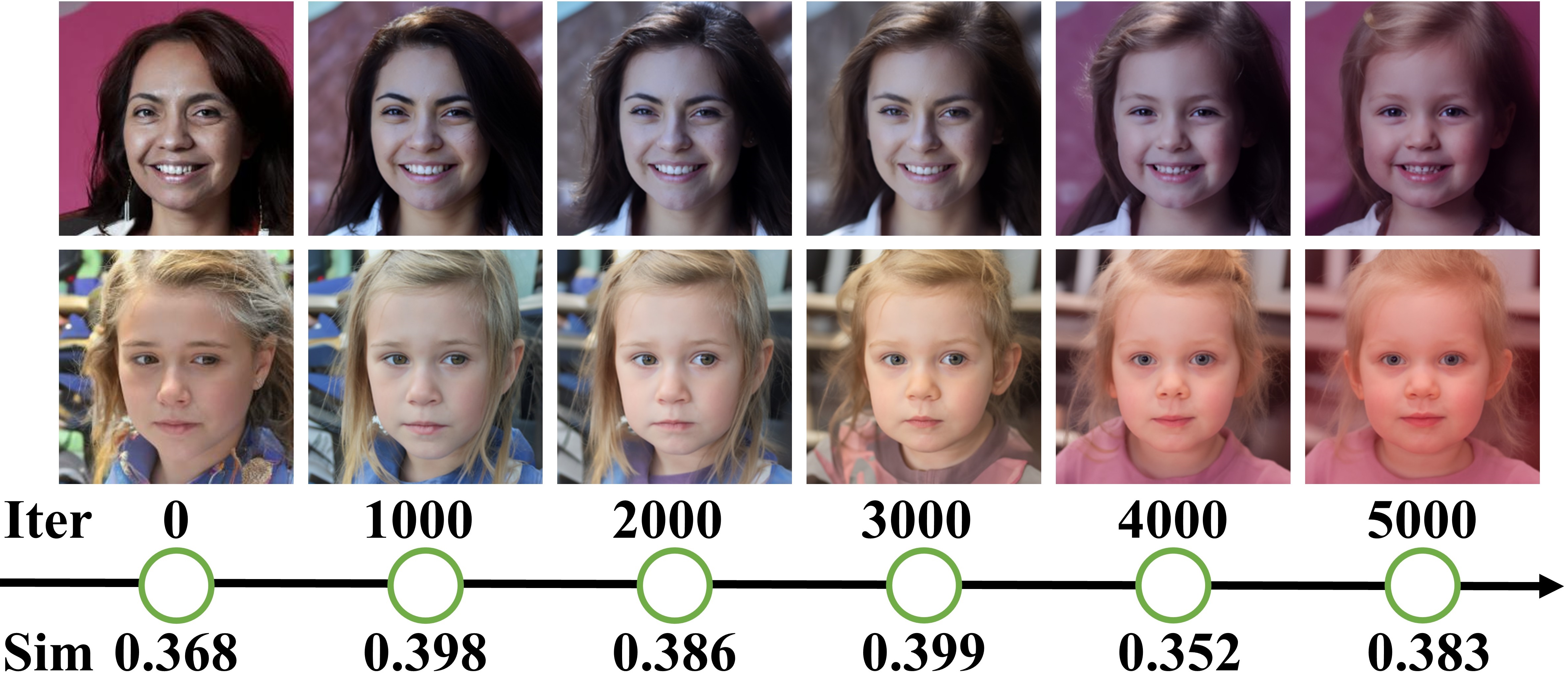}
    \caption{Samples synthesized from fixed noise inputs by DomainStudio on 10-shot FFHQ $\rightarrow$ Babies. DomainStudio keeps the relative pairwise distances during domain adaptation and achieves diverse results containing high-frequency details.}
    \label{degrade2}
\end{figure}

 \begin{figure}[t]
    \centering
    \includegraphics[width=1.0\linewidth]{ 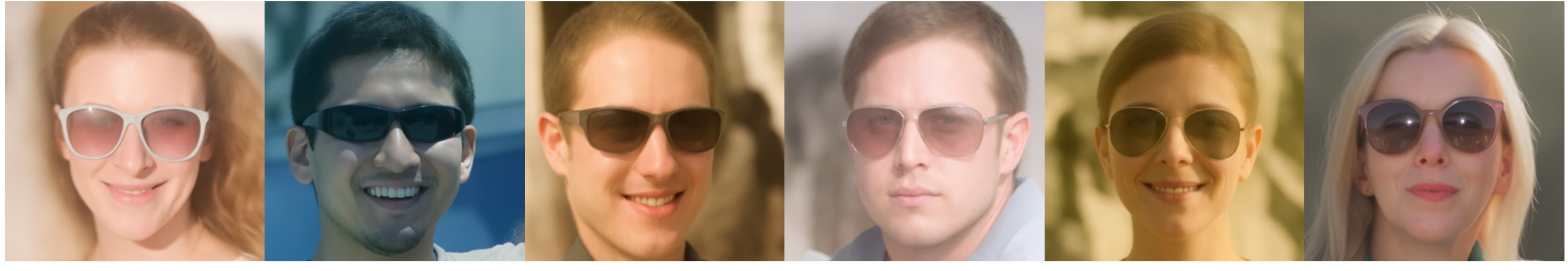}
    \caption{Samples synthesized by DomainStudio trained on 10-shot FFHQ $\rightarrow$ Sunglasses for 10000 iterations. Too many iterations lead to the replication of training samples and the degradation of generation quality and diversity.}
    \label{sunglass10000}
\end{figure}

\section{DDPM Adaptation Process Analysis}
\label{appendix_adaptation}
This paper mainly concentrates on the challenging 10-shot image generation tasks. When fine-tuning pre-trained DDPMs on target domains using limited data directly, too many iterations lead to overfitting and seriously degraded diversity. Fine-tuned models trained for about 10K iterations almost exclusively focus on replicating the training samples. Therefore, we train the directly fine-tuned DDPMs for 3K-4K iterations to adapt source models to target domains and maintain diversity. However, the directly fine-tuned DDPMs still generate coarse samples lacking details with reasonable iterations.

In Fig. \ref{amedeo}, we provide samples produced by DomainStudio models trained for different iterations on 10-shot FFHQ $\rightarrow$ Babies. We apply fixed noise inputs to different models for comparison. As the iterations increase, the styles of the generated images become closer to the training samples. Images synthesized from the same noise inputs as Fig. \ref{degrade} are included in red boxes. In addition, detailed evaluation of cosine similarity is added in Fig. \ref{degrade2}. The source samples are adapted to the target domain while keeping relatively stable cosine similarity. Compared with the directly fine-tuned DDPMs, DomainStudio has a stronger ability to maintain generation diversity and achieve realistic results containing rich details. Nonetheless, too many iterations still lead to details missing and the degradation of quality and diversity, as shown by the samples of the DomainStudio model trained on 10-shot FFHQ $\rightarrow$ Sunglasses for 10K iterations in Fig. \ref{sunglass10000}. Therefore, we recommend choosing suitable iterations for different adaptation setups (e.g., 4K-5K iterations for 10-shot FFHQ $\rightarrow$ Babies) to adapt the pre-trained models to target domains naturally and guarantee the high quality and great diversity of generated samples. 

 \begin{figure}[t]
    \centering
    \includegraphics[width=1.0\linewidth]{ 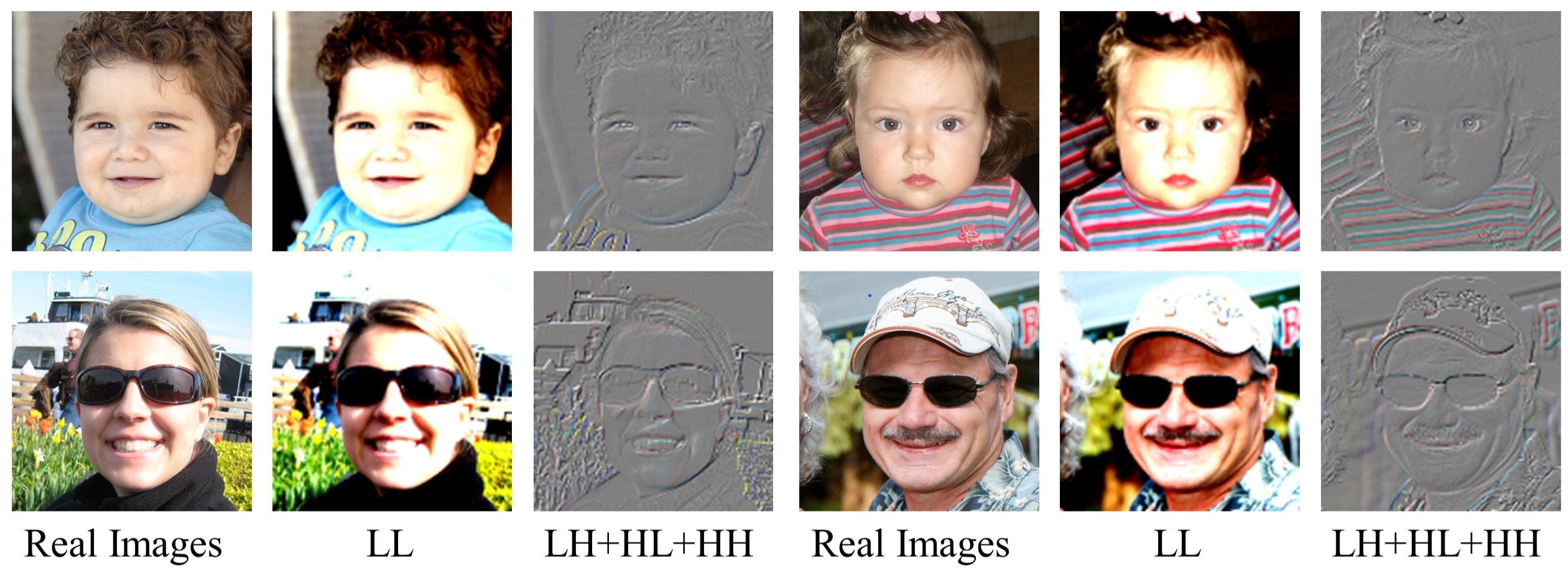}
    \caption{Visualization of the low and high-frequency components obtained with Haar wavelet transformation using images from Babies and Sunglasses as examples. LL represents the low-frequency components, and LH+HL+HH represents the sum of the high-frequency components.}
    \label{wavelet}
\end{figure}

\begin{figure*}[tbp]
    \centering
    \includegraphics[width=1.0\linewidth]{ 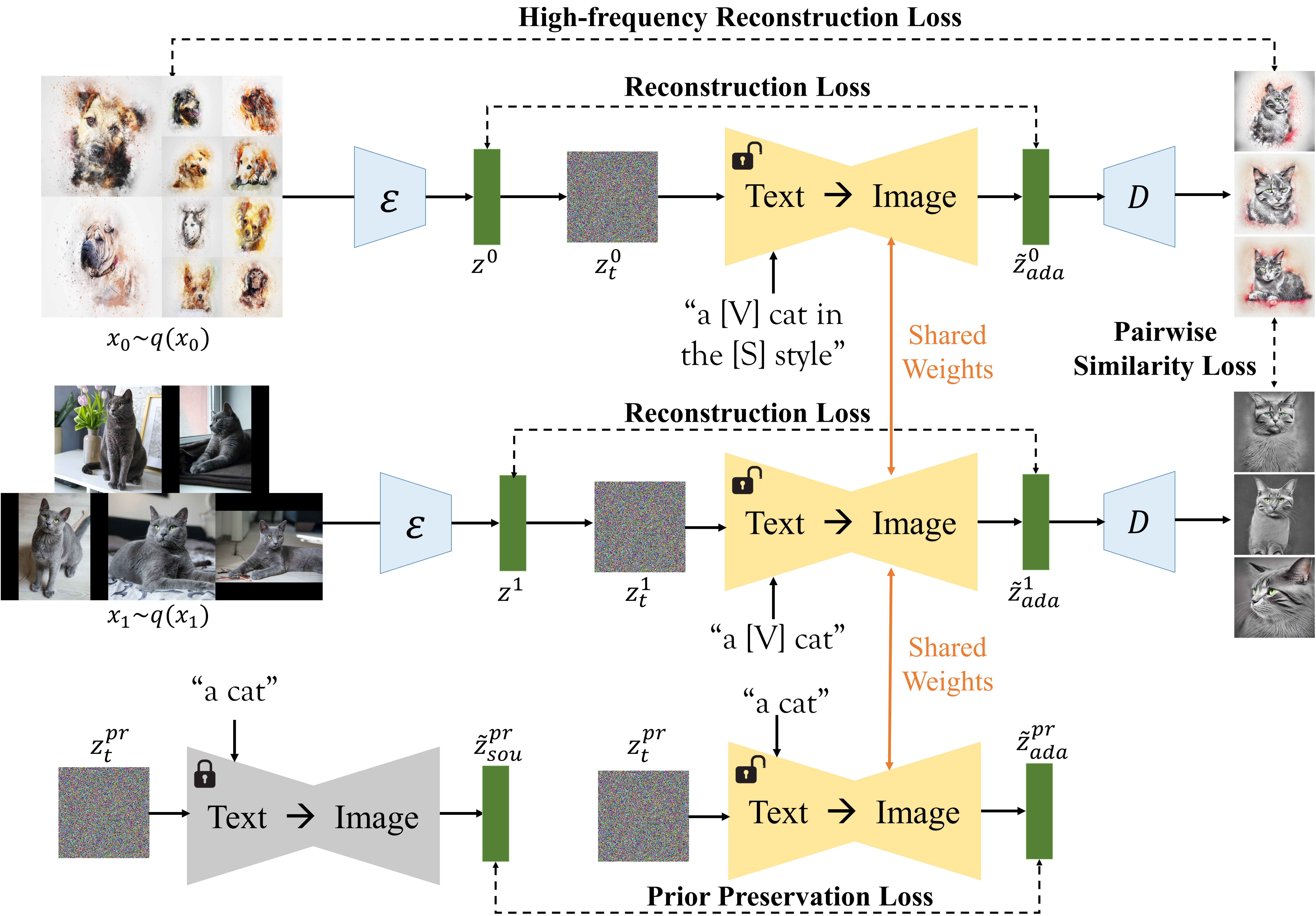}
    \caption{\textbf{Overview of the one-stage personalization of DomainStudio.} We combine DomainStudio with DreamBooth to achieve personalized domain-driven image generation.}
    \label{personal}
\end{figure*}

\begin{figure*}[tbp]
    \centering
    \includegraphics[width=1.0\linewidth]{ 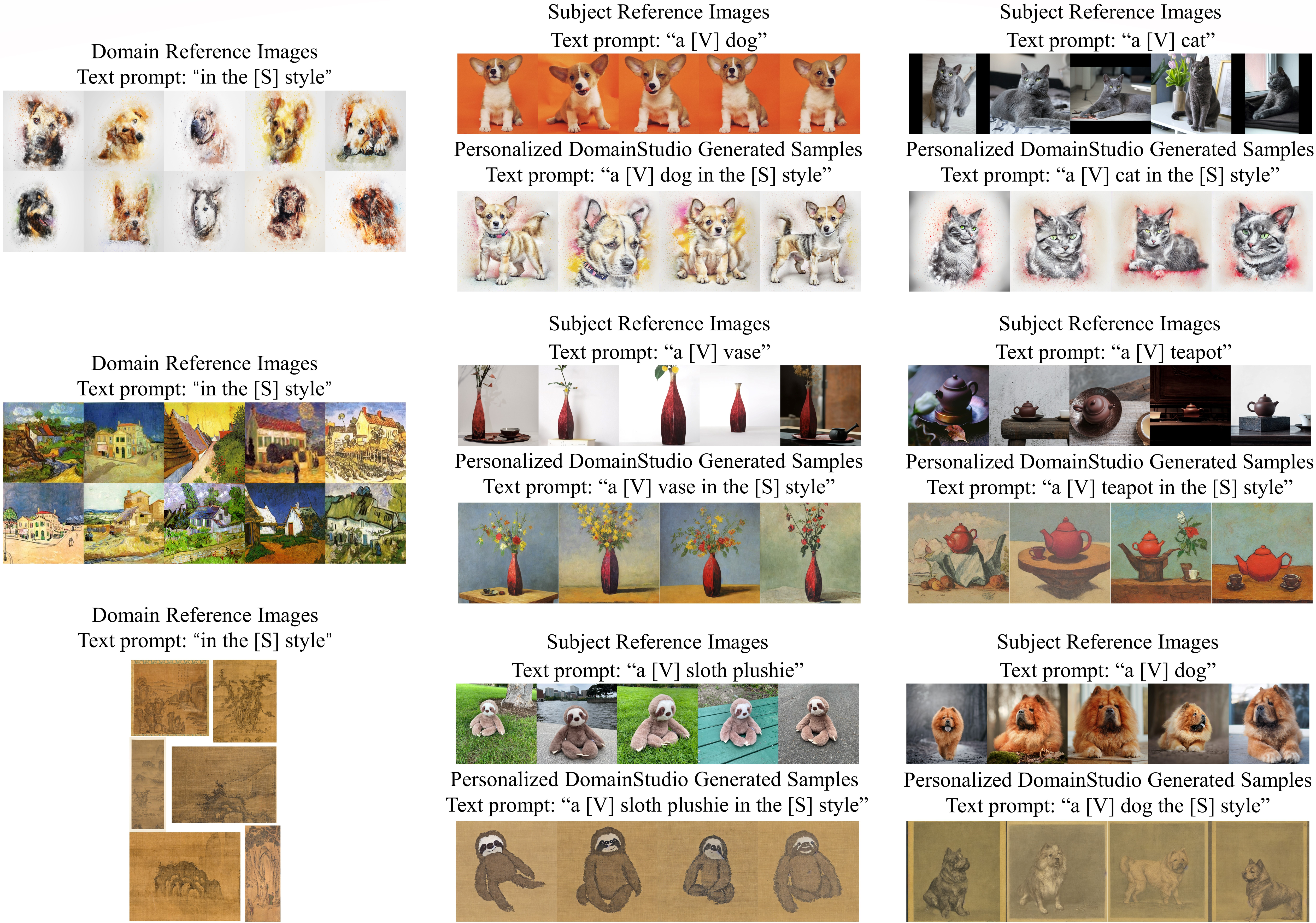}
    \caption{Image samples produced through the personalization of DomainStudio.}
    \label{personal_result}
\end{figure*}

\begin{figure*}[t]
    \centering
    \includegraphics[width=1.0\linewidth]{ 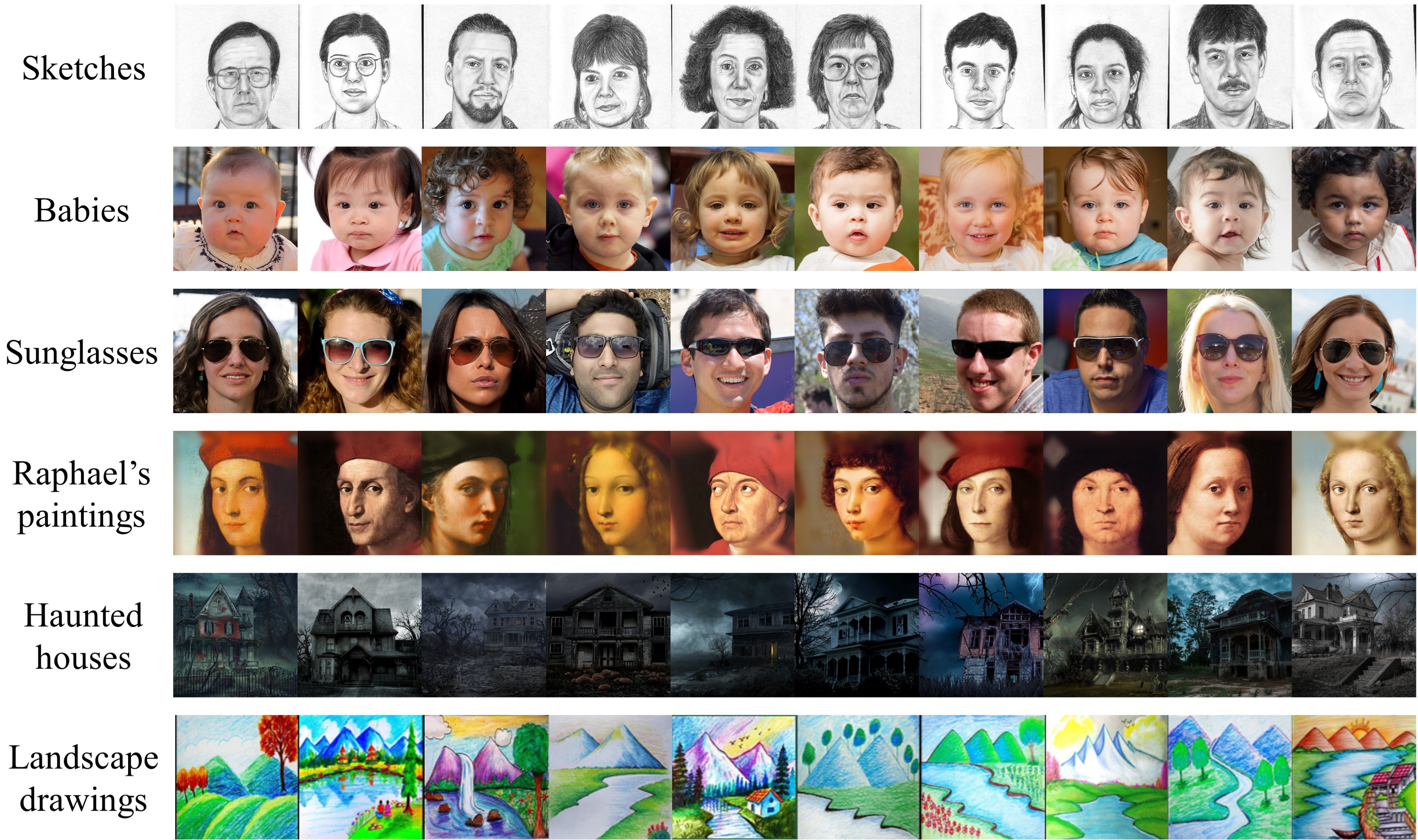}
    \caption{All the 10-shot datasets used in this paper, including 5 target domains corresponding to FFHQ and 2 target domains corresponding to LSUN Church.}
    \label{datasets}
\end{figure*}

\section{Harr Wavelet Transformation Examples}
Fig. \ref{wavelet} visualizes several examples of Haar wavelet transformation. The low-frequency components LL contain the fundamental structures of images. High-frequency components including LH, HL, and HH contain rich details like contours and edges in images.

\begin{figure*}[t]
    \centering
    \includegraphics[width=1.0\linewidth]{ 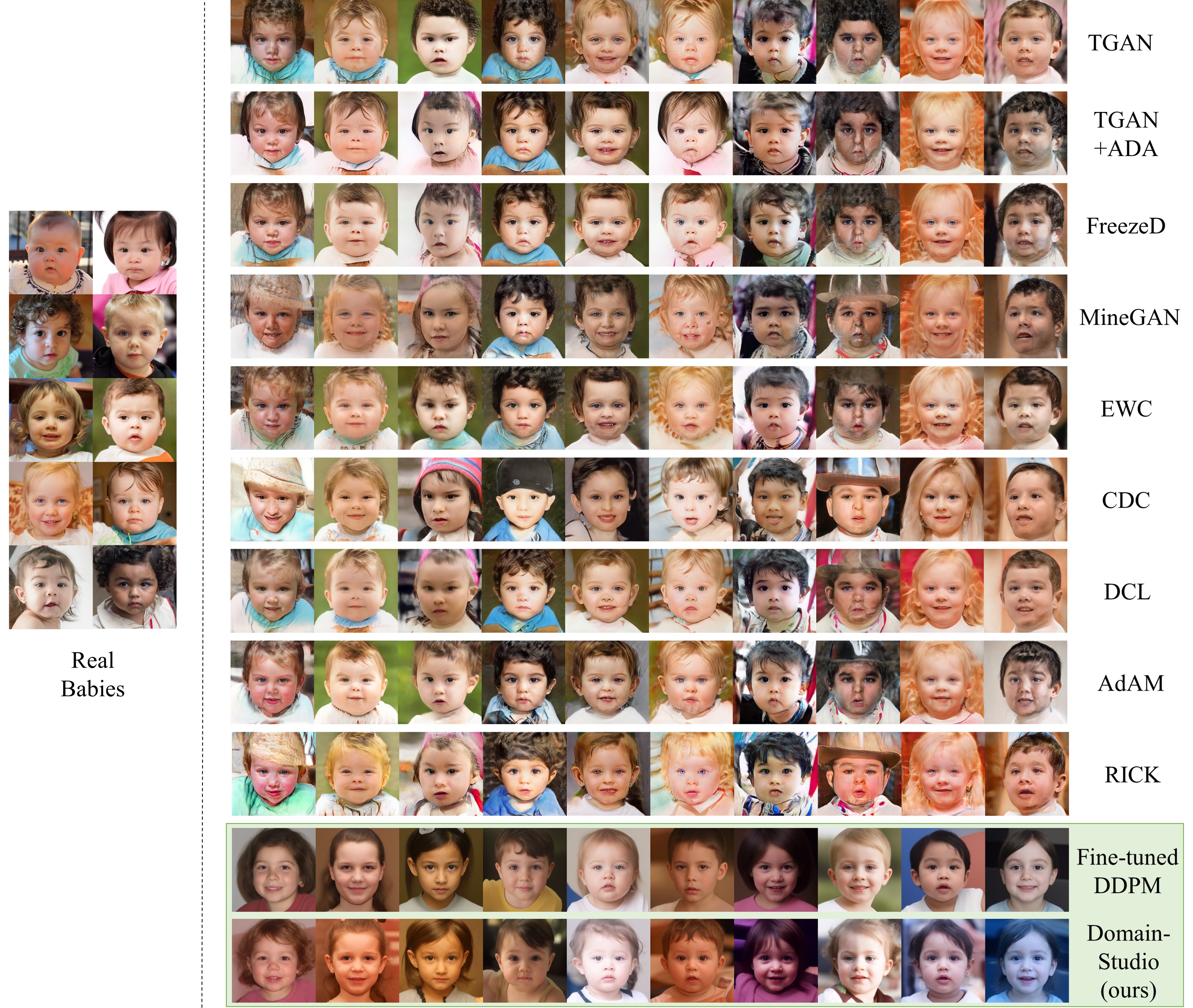}
    \caption{10-shot image generation samples on FFHQ $\rightarrow$ Babies. All the samples of GAN-based approaches are synthesized from fixed noise inputs (rows 1-9). Samples of the directly fine-tuned DDPM and DomainStudio are synthesized from fixed noise inputs as well (rows 10-11). }
    \label{babies}
\end{figure*}

\section{Personalization of DomainStudio}
\label{appendix_personal}
DomainStudio is designed to realize domain-driven generation, which differs from modern subject-driven approaches like DreamBooth \cite{ruiz2022dreambooth} and Textual Inversion \cite{gal2022textual}. In this section, we further explore the personalization of DomainStudio to satisfy both domain-driven and subject-driven requests. Given two sets of images as reference for the target subject and domain, we design two training strategies to realize the personalization of DomainStudio.

\textbf{One-Stage Approach} As illustrated in Fig. \ref{personal}, the proposed DomainStudio can be combined with DreamBooth to personalize domain-driven image generation. For example, we use text prompts: ``a cat", ``a [V] cat", and ``a [V] cat in the [S] style" corresponding to the source domain, personalized subject, and personalized subject in the target domain. Compared with the DomainStudio approach shown in Fig. \ref{pairwise2}, we add the reconstruction loss of personalized subjects and compute the pairwise similarity loss between personalized subjects and personalized subjects in the target domain.

\textbf{Two-Stage Approach} We can also divide the training process into two stages. Firstly, we follow DreamBooth to train the pre-trained text-to-image model to realize subject-driven generation. They we use the target subject as source domain and conduct DomainStudio to learn the knowledge of target domains while maintaining the personalized subject. 

In our experiments, we find that these two approaches achieve similar results. We provide several personalized domain-driven generation samples containing diverse subjects and styles in Fig. \ref{personal_result}. Our approach successfully adapt the personalized subject to target domains under the guidance of few-shot reference images. For instance, we adapt the reference dog and cat to the watercolor style (first row of Fig. \ref{personal_result}). Besides, we synthesize the reference vase and teapot in Van Gogh's style using 10-shot Van Gogh houses as domain reference (second row of Fig. \ref{personal_result}).



\section{Additional Visualized Samples}
\label{appendix_results}
We show all the 10-shot datasets used in this paper for few-shot image generation tasks in Fig. \ref{datasets}, including 5 target domains corresponding to the source domain FFHQ \cite{Karras_2020_CVPR} and 2 target domains corresponding to LSUN Church \cite{yu2015lsun}.

\begin{figure*}[t]
    \centering
    \includegraphics[width=1.0\linewidth]{ 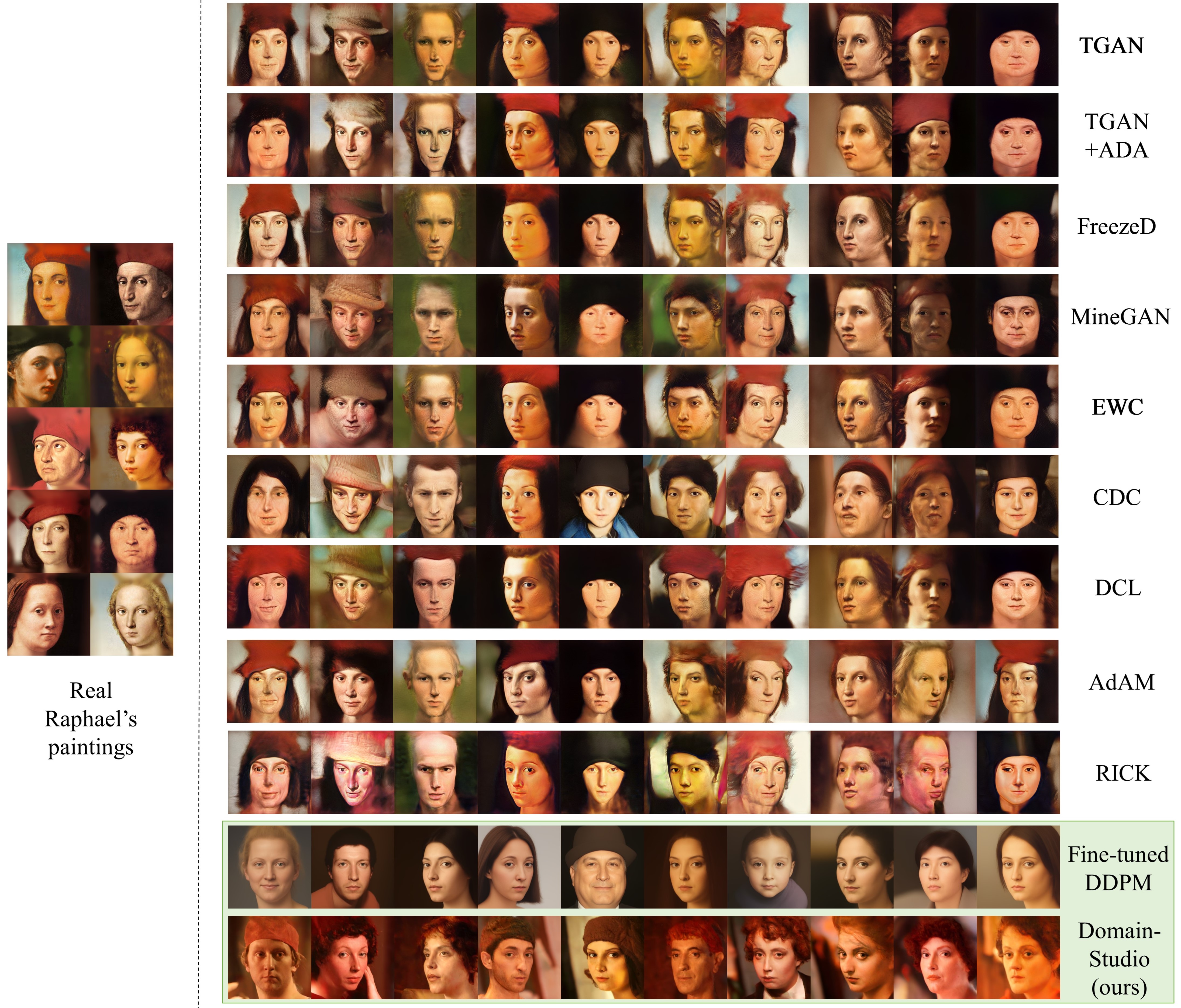}
    \caption{10-shot image generation samples on FFHQ $\rightarrow$ Raphael's paintings. All the samples of GAN-based approaches are synthesized from fixed noise inputs (rows 1-9). Samples of the directly fine-tuned DDPM and DomainStudio are synthesized from fixed noise inputs as well (rows 10-11). }
    \label{raphael}
\end{figure*}

\begin{figure*}[t]
    \centering
    \includegraphics[width=1.0\linewidth]{ 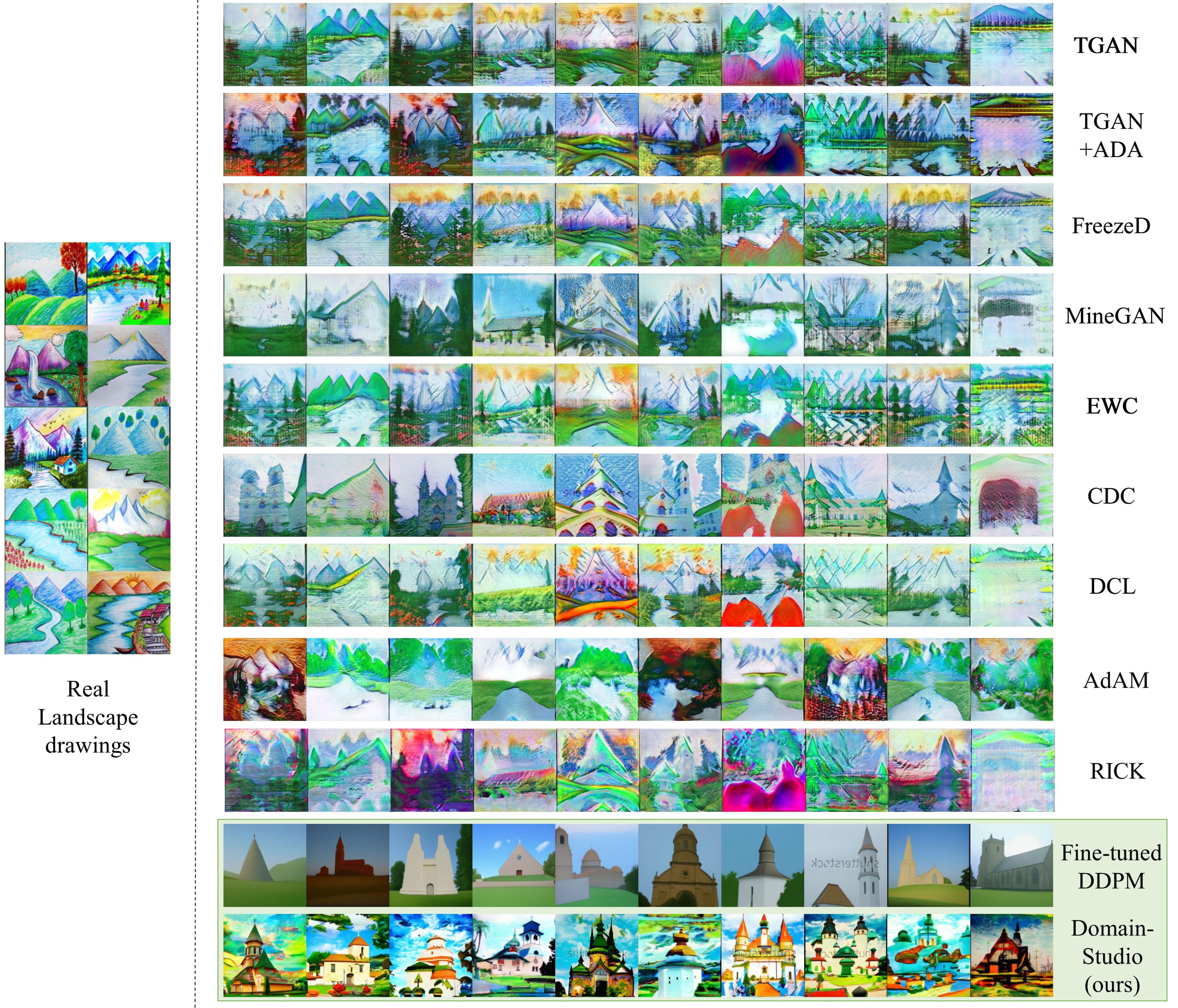}
    \caption{10-shot image generation samples on LSUN Church $\rightarrow$ Landscape drawings. All the samples of GAN-based approaches are synthesized from fixed noise inputs (rows 1-9). Samples of the directly fine-tuned DDPM and DomainStudio are synthesized from fixed noise inputs as well (rows 10-11). }
    \label{vangogh}
\end{figure*}

\begin{figure*}[t]
    \centering
    \includegraphics[width=1.0\linewidth]{ 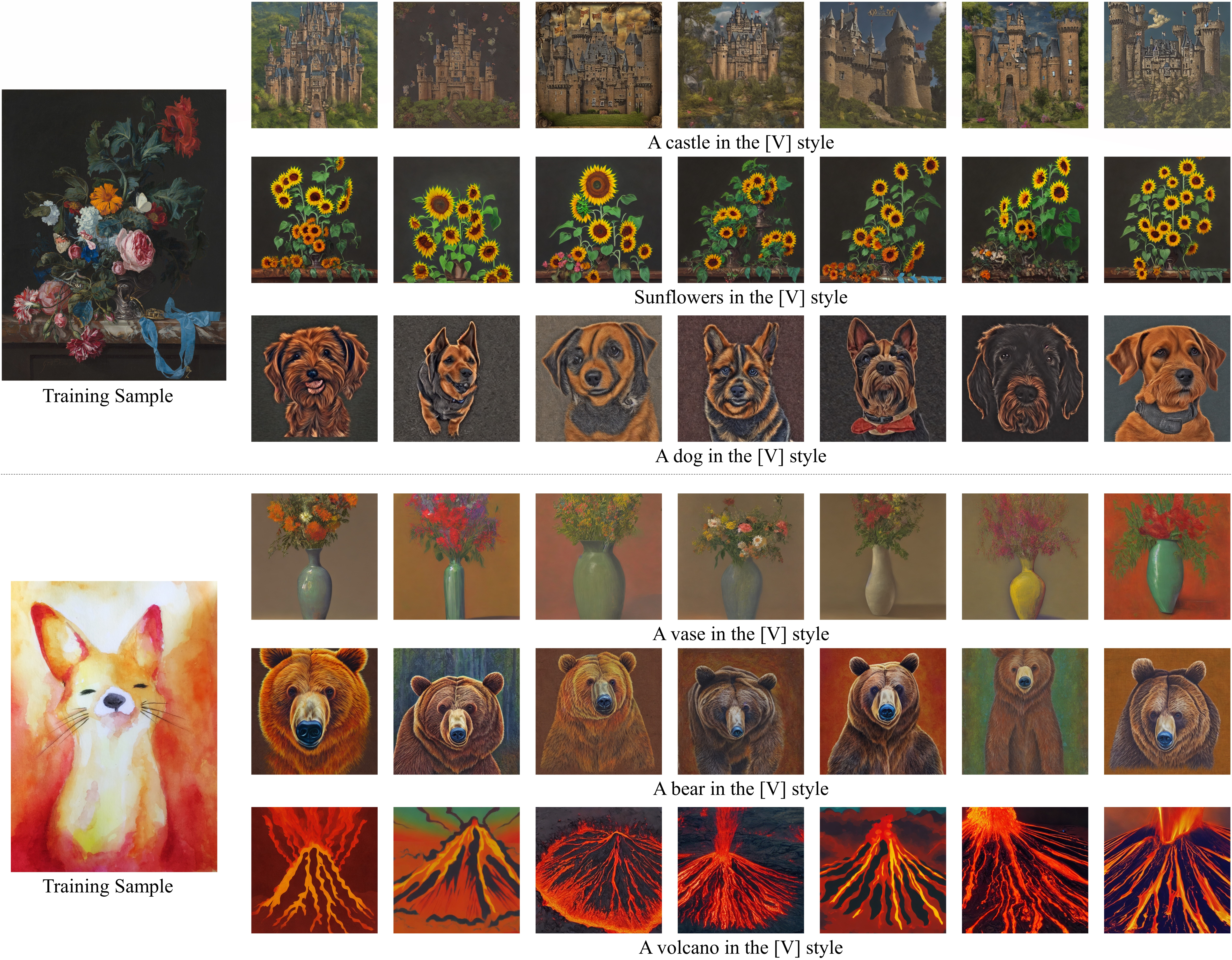}
    \caption{1-shot text-to-image generation samples of DomainStudio using different source domains.}
    \label{stable4}
\end{figure*}

Besides, we provide image generation samples of GAN-based baselines and DDPM-based approaches on 10-shot FFHQ $\rightarrow$ Babies, FFHQ $\rightarrow$ Raphael's paintings, and LSUN Church $\rightarrow$ Landscape drawings in Fig. \ref{babies}, \ref{raphael}, and \ref{vangogh} as supplements to Fig. \ref{sunglass}. We apply fixed noise inputs to GAN-based approaches and DDPM-based approaches, respectively. DDPMs are more stable and less vulnerable to overfitting than GANs. GANs easily overfit and tend to generate samples similar to training samples when training data is limited (see samples of TGAN \cite{wang2018transferring}). Directly fine-tuned DDPMs can still keep a measure of generation diversity under the same conditions. Besides,  DDPM-based approaches avoid the generation of blurs and artifacts. However, directly fine-tuned DDPMs tend to produce too smooth results lacking high-frequency details and still face diversity degradation. DomainStudio generates more realistic results containing richer high-frequency details than GAN-based baselines under all these adaptation setups.

DomainStudio performs apparently better when we want to keep the fundamental structures of source images and learn the styles of target domains (e.g., LSUN Church $\rightarrow$ Landscape drawings). As shown in Fig. \ref{vangogh}, GAN-based approaches fail to adapt real churches to the drawing style and produce samples containing too many blurs and artifacts. On the other hand, DomainStudio produces high-quality church drawings and preserves more detailed building structures.

Fig. \ref{stable4} shows results of DomainStudio on text-to-image generation using a single image as training data. It's hard to define the target domain accurately with a single image. We recommend using 4-10 images to realize diverse, high-quality, and stable domain-driven text-to-image generation.

\section{Computational Cost} 
The time cost of DDPMs and the proposed DomainStudio approach are listed in Table \ref{timecost}. DomainStudio costs $24.14\%$ more training time than the original DDPMs. DDPMs trained from scratch need about 40K iterations to achieve reasonable results, even if they can only replicate the training samples. DomainStudio utilizes related pre-trained models to accelerate convergence (about 3K-5K iterations) and significantly improve generation quality and diversity. Compared with directly fine-tuned DDPMs, DomainStudio is not overly time-consuming and achieves more realistic results.

\begin{table}[t]
    \centering
    \begin{tabular}{l|c}
        Approaches & Time Cost $/\ $ 1K Iterations   \\
        \hline
        DDPMs  &  29 min   \\
        DomainStudio (ours) &  36 min    
    \end{tabular} 
    \caption{The time cost of directly fine-tuned DDPMs (batch size 48) and DomainStudio (batch size 24) models trained for 1K iterations on $\times 8$ NVIDIA RTX A6000 GPUs.}
    \label{timecost}
\end{table}

\begin{table}[t]
    \centering
    \begin{tabular}{l|c}
        Approaches & Time Cost $/\ $ 1500 Iterations   \\
        \hline
        Textual Inversion \cite{gal2022textual}  &  33 min   \\
        DreamBooth \cite{ruiz2022dreambooth} & 8 min \\
        DomainStudio (ours) &  21 min \\    
    \end{tabular} 
    \caption{The time cost of fine-tuned Stable Diffusion trained for 1.5K iterations on a single NVIDIA RTX A6000 GPU.}
    \label{timecost2}
\end{table}

The time cost of DomainStudio applied to text-to-image generation is shown in Table \ref{timecost2}. DomainStudio is more efficient than Textual Inversion \cite{gal2022textual}. It is more time-consuming than DreamBooth \cite{ruiz2022dreambooth} since we add the computation of pairwise distance maintenance and high-frequency details enhancement. However, it is still acceptable to fine-tune a model using about 20 minutes on a single GPU.


\end{document}